\documentclass[12pt]{article}

\usepackage{amsmath, amsthm, amsfonts, amssymb}
\usepackage{graphicx,psfrag,epsf}
\usepackage{enumerate}
\usepackage{natbib}
\usepackage{bbm}
\usepackage{hyperref}
\usepackage{cleveref}
\usepackage{bm}
\usepackage{algorithm}
\usepackage{setspace}
\usepackage{longtable}
\usepackage{comment}
\usepackage{algpseudocode}
\usepackage{booktabs}
\usepackage{subcaption}
\usepackage{multicol}
\usepackage{multirow}
\usepackage{adjustbox}
\usepackage[table]{xcolor}
\usepackage{url} 
\usepackage{ragged2e}
\usepackage{amsbsy}
\usepackage{authblk}
\usepackage{mathtools}
\usepackage{tikz}
\usetikzlibrary{arrows.meta,calc}
\usetikzlibrary{decorations.pathreplacing}
\usepackage{natbib}
\usepackage{arydshln}
\newcommand{\blind}{1}

\theoremstyle{plain}
\newtheorem{theorem}{Theorem}[]
\newtheorem{corollary}{Corollary}[theorem]
\newtheorem{assumption}{Assumption}[]
\newtheorem{lemma}{Lemma}[section]
\theoremstyle{definition}

\theoremstyle{remark}

\def\R{\mathbb{R}}

\def\bY{\bm{Y}}
\def\bX{\bm{X}}
\def\bC{\bm{C}}
\def\bV{\bm{V}}

\def\bW{\bm{W}}

\newcommand{\deltaeta}{\bm{\delta}_{\bm{\eta}}}
\newcommand{\Q}{\bm{Q}}

\renewcommand{\vec}[1]{\mathbf{1}}
\DeclareMathOperator*{\argmin}{arg\min}

\newcommand{\E}{\mathbb{E}}

\newcommand{\boldeta}{\bm\eta}

\begin{document}

\def\spacingset#1{\renewcommand{\baselinestretch}%
{#1}\small\normalsize} \spacingset{1}


\if1\blind
{
  \title{\bf Hierarchical Projection for Adaptive Knowledge Transfer}

    \author[1]{Samhita Pal}
\author[2,\footnote{Corresponding author. Email address: \texttt{tg2880@cumc.columbia.edu}}]{Tian Gu}
\affil[1]{\small Department of Biostatistics, Vanderbilt University Medical Center, Nashville, TN}
\affil[2]{Department of Biostatistics, Columbia University Mailman School of Public Health, New York, NY}
\date{}

  \maketitle
} \fi

\if0\blind
{
  \bigskip
  \bigskip
  \bigskip
  \begin{center}
    {\LARGE\bf Hierarchical Projection for Adaptive Knowledge Transfer}
\end{center}
  \medskip
} \fi

\bigskip
\begin{abstract}
Transfer learning aims to improve estimation in a target population with limited data by borrowing information from related external datasets, referred to as sources. However, heterogeneity in target-source relevance and predictor-outcome relationships can lead to negative transfer, where model performance is worse than using target data alone, particularly in high-dimensional settings with partially overlapping features. We propose Projection Transfer Learning (ProjectionTL), a Bayesian framework that integrates hierarchical modeling with adaptive posterior projection to enable selective knowledge transfer at both the source and feature levels. In the first stage, a source-guided hierarchical prior adaptively combines information from multiple sources using data-driven weights that capture their global alignment with the target. In the second stage, a weighted posterior-projection procedure induces sparsity by preferentially retaining features supported by stronger agreement between source and target signals. This two-stage framework distinguishes global source relevance from coordinate-specific transferability, allowing informative components to be retained while reducing the influence of heterogeneous or non-transferable signals. We establish theoretical guarantees for posterior contraction and model selection consistency under high-dimensional regularity conditions, without requiring all sources to be informative. Simulation studies and applications to Alzheimer's disease neuroimaging data demonstrate improved estimation accuracy, variable selection, and predictive performance compared with others, including settings with heterogeneous sources and block-wise missing modalities.
\end{abstract}

\noindent%
\vfill

\newpage
\spacingset{1.8} 
\section{Introduction}\label{sec:intro}
In many scientific domains, it is increasingly common to have multiple heterogeneous datasets that partially overlap in features or modalities, yet share a common prediction task. These datasets may vary in quality, scale, and completeness, leading to challenges in jointly learning from them without being misled by uninformative or misaligned sources. Our objective is to develop a principled transfer learning framework that selectively borrows information from multiple sources to improve estimation and prediction in a target dataset, while maintaining robustness to heterogeneity and block-wise missingness. 

Transfer learning in high-dimensional linear regression has gained popularity with \cite{bastani2021predicting}, who considered a single source. This was extended to multiple sources via Trans-LASSO by \cite{li2022transfer} and to generalized linear models and others soon after \citep[e.g.,][]{li2024estimation, tian2023transfer, cai2024transfer, gu2025connections}. These methods have several challenges. Firstly, some of them require individual-level data from the source cohorts, which is often infeasible nowadays owing to data-sharing constraints, thereby leading to federated learning algorithms or approaches that require only summary-level source information \citep[e.g.,][]{li2023targeting, gu2023commute, lu2024enhancing}. Secondly, these methods assume complete overlap of the feature space between the target and all sources. Consequently, they may suffer from negative transfer when only subsets of features are shared or when sources exhibit heterogeneous, coordinate-specific deviations from the target. Some works address this limitation of heterogeneous and non-overlapping sources \citep[e.g.,][]{duan2022heterogeneity, zhao2023heterogeneous, chang2024heterogeneous, zhao2023heterogeneous, pal2026improving, asiaee2026improving, ghosh2026demographic, asiaee2026b}. In many contemporary applications, these challenges arise naturally. For example, in biomedical studies, multi-center cohorts often collect overlapping but non-identical modalities, such as imaging, genomics, and clinical variables, with substantial variation in measurement protocols and population composition. Similarly, in policy evaluation and real-world evidence generation, historical datasets may differ in covariate availability or data quality, yet contain partially relevant signals for a current target population. These settings require methods that can adaptively borrow information at a finer granularity than the dataset level, while remaining robust to structured heterogeneity and missingness.

Most of the above discussed approaches rely on $\ell_1$ or $\ell_2$-norm based distance metrics that capture the amount of agreement between the sources and the target coefficient estimates. This is ineffective, especially when sources and target models' parameters differ in scale or structure. To address this, \cite{gu2024robust} devised an angle-based distance metric and later \cite{lai2026bayesian} developed a Bayesian transfer learning algorithm called TRADER that uses a similar idea to transfer knowledge from sources along with the horseshoe prior for sparsity \citep{lai2026bayesian}. Although these methods can borrow adaptively from different sources with different strengths of similarity with the target, they also assume complete overlap of covariates and complete match of signals. \citep{zhang2024covariate}, on the other hand, proposed CONCERT, a Bayesian high-dimensional transfer-learning method that uses covariate-specific spike-and-slab priors to decide which variables should borrow information from which sources. Although CONCERT allows for covariate-level partial transfer, it does not explicitly model source-level heterogeneity through adaptive source weights like TRADER, so that once a covariate is deemed transferable, the method does not separately learn how strongly each auxiliary source should contribute.

Related literature addresses multi-modal data integration when entire blocks of covariates are missing across subjects or studies. For high-dimensional regression, DISCOM \citep{yu2020optimal} and its multi-response extension \citep{wang2024multiresponse} exploit available covariance information to estimate sparse regression models without explicit imputation. Other approaches use block-wise
imputation and bias-corrected estimating equations to support estimation and inference under structured missingness \citep{xue2025statistical}. More recently,
\cite{li2024adaptive} proposed DEFUSE, which uses data-adaptive projection to improve estimation efficiency when block-wise missing covariates coexist with partially
observed outcomes. Related developments also consider domain adaptation under distribution shift and block-wise missing modalities \citep{wang2026multimodal}.
While these methods address important challenges in integrating incomplete multi-modal data, our focus is distinct: we seek to adaptively borrow information from
heterogeneous source models to improve estimation in a data-limited target population, combining source-level weighting with coordinate-level selective transfer under partially overlapping covariates.

We propose Projection Transfer Learning (ProjectionTL), a Bayesian framework that integrates adaptive borrowing across heterogeneous sources with selective refinement at the covariate level. ProjectionTL combines two complementary stages: a hierarchical prior performs source-level borrowing by weighting source models according to their global alignment with the target, while a posterior-projection step performs component-level refinement by inducing sparsity and preferentially retaining coefficients supported by stronger source--target agreement. The projection step builds on recent developments in projection-posterior methods, which map dense posterior draws onto sparse representations to facilitate variable selection and improve interpretability \citep{wang2023posterior,wang2023coverage,chakraborty2022rates,pal2025projection,pal2024group,pal2024bayesian,gili2026semiparametric}. Unlike existing projection-posterior approaches that primarily focus on single-dataset settings, ProjectionTL incorporates external source information into posterior learning and adaptive sparsification. To accommodate partially overlapping feature spaces, we further introduce source-specific calibration that accounts for differences between reduced-model source coefficients and target full-model coefficients.

Our contributions are fourfold. First, we develop a unified hierarchical projection framework that combines source-level weighting and component-level adaptive sparsity for selective knowledge transfer. Second, we introduce target-informed source weights with a no-transfer option and extend the framework to partially overlapping covariates through source-specific calibration, requiring only coefficient estimates rather than individual-level source data. Third, we establish theoretical guarantees for posterior contraction, model selection consistency, and uncertainty quantification in high-dimensional settings. In particular, our analysis characterizes the posterior behavior of the adaptively learned source weights, establishes a safe contraction rate without requiring informative sources, and identifies conditions under which source information improves the contraction bound. Finally, simulations and applications to Alzheimer's disease neuroimaging data demonstrate improved estimation accuracy, variable selection, and predictive performance under source heterogeneity and block-wise missingness.

The rest of the paper is organized as follows. In Section \ref{sec:method}, we introduce the proposed ProjectionTL framework and describe its hierarchical prior and projection components. Section \ref{sec:theory} presents theoretical results, Section \ref{sec:simu} evaluates performance through extensive simulation studies under varying degrees of overlap and heterogeneity. Section \ref{sec:real_data} illustrates the method on real-world data from multi-phase Alzheimer’s disease studies. Section \ref{sec:discussion} concludes with a discussion of limitations and future directions.


\vspace{-1em}
\section{Methodology}\label{sec:method}


\subsection{Problem Setup and Notation}
We observe a target dataset $(\boldsymbol{X},\boldsymbol{Y})$, where
$\boldsymbol{X}\in\mathbb{R}^{n_0\times p}$ and $\boldsymbol{Y}\in\mathbb{R}^{n_0}$,
along with $K$ auxiliary (source) datasets $\{(\bm X_k, \bm Y_k)\}_{k=1}^K$, each providing a fitted coefficient vector
$\widehat{\bm \omega}_k=(\widehat{\omega}_{k1},\ldots,\widehat{\omega}_{kp})^\mathrm{T},~k=1,\ldots,K$.
We want to estimate a {sparse} target coefficient vector
$\boldsymbol{\beta}\in\mathbb{R}^p$ from modeling the linear relationship $\boldsymbol{Y}|\boldsymbol{X}$, but may yield unsatisfactory performance due to the small target sample size $n_0$. Given that $\widehat{\bm \omega}_k$'s may share certain similarity with $\boldsymbol{\beta}$, we want an improved model by selectively borrowing information from these source estimates while preventing negative transfer from incorporating misaligned sources, leading to worse performance than using only the target.

\subsection{Hierarchical Prior from Sources (Vector-Level Borrowing)}
\begin{figure}[h]
    \centering
    \includegraphics[width=0.9\linewidth]{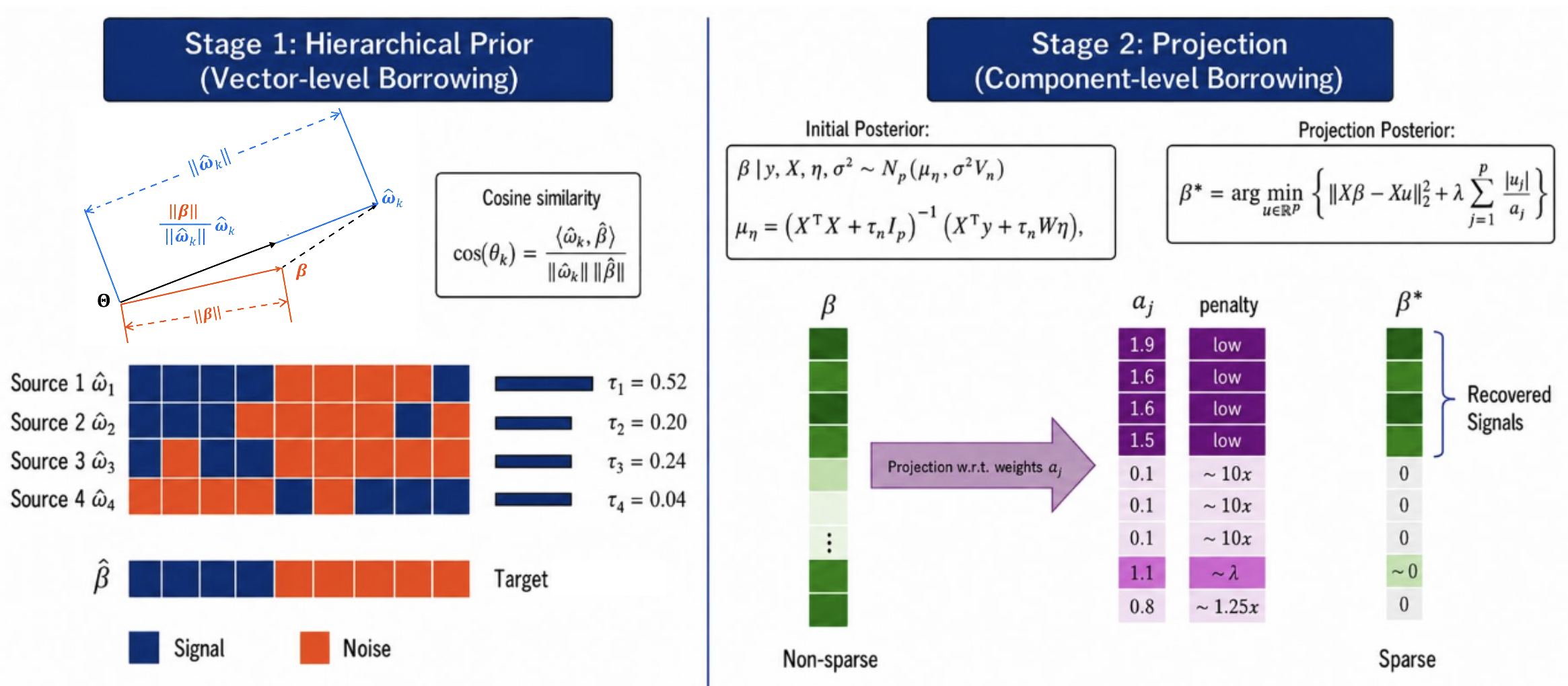}
    \caption{Schematic diagram of the proposed two-stage ProjectionTL.}
    \label{fig:diagram_full}
\end{figure}

As illustrated in the left panel of Figure~\ref{fig:diagram_full}, we begin by constructing an informative prior for the target parameter by borrowing information from multiple available sources. Ideally, we want sources that are more similar to the target to contribute more, while less relevant sources contribute little or nothing. Since source estimates may be estimated on a different scale than the target, naively averaging them may mislead the target estimate in the wrong direction \citep{gu2024robust,lai2026bayesian}. Therefore, we first rescaled the $k$-th source estimate to the same scale as the target via the scaling factor ${\|\widehat{\bm\beta}\|}/{\|\widehat{\bm\omega}_k\|}$. Then, for each target coefficient $\beta_j$, we construct a prior mean $\mu_j$ by taking a rescaled weighted average of the corresponding coefficients estimated from $K$ source models, $\widehat{\omega}_{kj}$:
\begin{equation}\label{eq:prior-mean}
  \mu_j \;=\; \sum_{k=1}^K \eta_k \cdot \frac{\|\widehat{\bm\beta}\|}{\|\widehat{\bm\omega}_k\|}\cdot\widehat{\omega}_{kj} + \eta_{K+1} \cdot 0, \qquad j=1,\ldots,p,
\end{equation} 
where $\widehat{\boldsymbol{\beta}}$ is the estimated coefficient vector using the target dataset. A natural alternative is to split the target sample and use a validation-based pilot estimate to guide the prior-centering mechanism. However, this can substantially reduce sample efficiency and introduce instability when the target sample size is limited. Therefore, we use the full target sample to construct $\widehat{\boldsymbol{\beta}}$, treating it as an empirical-Bayes hyperparameter construction step rather than direct reuse of the target likelihood. Similar strategies are common in Bayesian information-borrowing and transfer-learning methods \citep{jin2021unit,zhang2023unit,lai2026bayesian}. Additional discussion is provided in Supplement C.

The weights $\eta_k$ determine how much each source contributes to $\mu_j$, assigning larger values to sources that align well with the target signal and smaller ones to unhelpful sources. On the other hand, $\eta_{K+1}$ denotes the prior mean share when none of the sources are aligned with the target estimates. Since it multiplies to zero, in the case of no source alignment, the prior falls back to a mean-zero non-informative normal distribution. The vector of weights $\boldsymbol{\eta}=(\eta_1,\ldots,\eta_K, \eta_{K+1})^\mathrm{T}$ lies on the probability simplex, ensuring that $\mu_j$ is a convex combination of the source coefficients, which keeps the prior mean interpretable as the relative contribution (fraction) of each source. Specifically, the weights are learned through a Dirichlet prior whose concentration parameters $\theta_k$ quantify the alignment of each source estimate $\widehat{\boldsymbol{\omega}}_k$ with $\widehat{\boldsymbol{\beta}}$ via the cosine similarity: that is, $\boldsymbol{\eta} \sim \mathrm{Dirichlet}(\theta_1,\ldots,\theta_K,\zeta), 
  \text{ with }
  \theta_k \;=\; 
  {\widehat{\bm \omega}_k^\mathrm{T} \widehat{\boldsymbol{\beta}}}/
       ({\|\widehat{\bm \omega}_k\|\cdot\|\widehat{\boldsymbol{\beta}}\|}),$
\normalsize where the $(K\!+\!1)^\textnormal{th}$ component with hyperparameter $\zeta>0$ corresponds to a ``no-borrow'' option (i.e., target-only estimation) that provides additional shrinkage toward zero when sources are weak or misaligned. $\theta_k$ provides a scale-free measure of agreement that emphasizes directional alignment rather than magnitude. Sources with stronger directional alignment with the target, as measured by cosine similarity, receive larger $\theta_k$ and are thus more likely to obtain higher posterior weights $\eta_k$, whereas poorly aligned sources are downweighted. To ensure valid Dirichlet concentration parameters, we impose a small positive lower bound on the cosine-similarity scores, so that negatively aligned sources receive minimal prior concentration without altering their coefficient estimates. 
We set $\zeta = 1$ as the default. 
We then perform vector-level source selection through a simple conjugate normal prior conditional on $(\mu_j,\sigma^2)$. 
Specifically, we construct the hierarchy $\beta_j \mid \mu_j,\sigma^2 \sim \mathcal{N}(\mu_j, \tau_n^{-1}\sigma^2)
  \text{ with }
  \sigma^2 \sim \text{Inverse-Gamma}(\nu,\nu),$
which performs source-level borrowing by adaptively weighting source estimates according to their agreement with the target. 
The subsequent projection step operates at the component level, refining borrowing by selecting which specific variables within each source contribute.


\subsection{Posterior Projection with Adaptive Sparsity (Component-Level Refinement)}
After Stage 1, we have dense posterior draws of $\bm\beta$. While the hierarchical prior identifies which source datasets are globally informative, it does not assume that every coefficient from an informative source is transferable. To project this dense posterior onto a sparser, more interpretable model, we therefore refine the posterior obtained from Stage 1 through an adaptive projection that selectively retains coefficients supported by consistent evidence across sources (right panel of Figure~\ref{fig:diagram_full}). 

The conditional posterior of the target parameter obtained from Stage 1 is $\bm\beta \mid \bm y,\bm X,\bm\eta,\sigma^2
\sim
N_p\!\left(\bm\mu_{\eta},\sigma^2 \bm V_n\right)$
with $
\bm V_n =
\left(\bm X^\mathrm{T} \bm X + \tau_n \bm I_p\right)^{-1}$
and $\bm\mu_{\eta}
=
\bm V_n
\left(\bm X^\mathrm{T} \bm y + \tau_n \bm W\bm\eta\right),$ {where $\bm W \in \mathbb{R}^{p \times (K+1)}$ is the rescaled source
coefficient matrix with $(j,k)$-th entry
$W_{jk} = ({\|\widehat{\bm\beta}\|}/{\|\widehat{\bm\omega}_k\|})\; \widehat{\omega}_{kj}$ for $k = 1,\ldots,K$
and $W_{j,K+1} = 0$ for all $j$, so that $\bm W\bm\eta$ recovers
the source-guided prior mean $\bm\mu_j$ in
Equation~\eqref{eq:prior-mean}.} We then project each posterior draw $\bm\beta$ onto a weighted $\ell_1$-penalized surface using the target design:
{\small\begin{equation}\label{eq:projection}
  \boldsymbol{\beta}^\ast 
  \;=\; \arg\min_{\boldsymbol{u}\in\mathbb{R}^p}
  \Big\{
     \|\boldsymbol{X}\boldsymbol{\beta}-\boldsymbol{X}\boldsymbol{u}\|_2^2
     \;+\;
     \lambda \sum_{j=1}^p \frac{|u_j|}{a_j}
  \Big\}.
\end{equation}}
The adaptive penalty weight $a_j$ for coordinate $j$ is defined as $a_j =
    \sum_{k=1}^K 
      \exp\{-(\widehat{\beta}_j - W_{jk})^2\},$
which aggregates the agreement between the target estimate and all source estimates at that coordinate. Each exponential term contributes a value between zero and one, with larger values indicating stronger agreement between the target and a particular source. Consequently, $a_j$ measures the cumulative evidence supporting transfer of coefficient $j$ across all sources. Coordinates exhibiting broad agreement across multiple sources receive larger values of $a_j$ and therefore smaller effective penalties $1/a_j$, making them more likely to be retained after projection. In contrast, if most or all sources disagree with the target estimate, $a_j$ becomes small, increasing the effective penalty and shrinking the corresponding coefficient toward zero. The tuning parameter $\lambda\!>\!0$ controls the overall sparsity level and can be chosen by cross-validation or by targeting a projection loss tolerance. 

To obtain the final estimator, we use the Median Probability Model (MPM). Let $\pi_j^\ast=\Pi_{\mathrm{ProjectionTL}}(\beta_j^\ast\neq 0
\mid\mathcal D_0,\ldots,\mathcal D_K)$ denote the posterior inclusion
probability of coordinate $j$. The final estimator is defined as
\begin{equation}\label{eq:final-estimator}
\widehat\beta_j^{\mathrm{MPM}} = \mathbb E_{\mathrm{ProjectionTL}}
\left(\beta_j^\ast\mid\mathcal D_0,\ldots,\mathcal D_K\right)
\mathbb I\left(\pi_j^\ast>0.5\right),
\qquad j=1,\ldots,p.
\end{equation}
Thus, coefficients selected in more than half of the projected posterior draws are retained and estimated by their projected posterior means, while the remaining coefficients are set to zero. This way, the projected draws $\boldsymbol{\beta}^\ast$ are aggregated across posterior samples to obtain a sparse point estimate together with projection-based inclusion probabilities. 

The proposed framework thus separates transfer learning into two complementary stages. The hierarchical prior with source-guided mean via equation~\eqref{eq:prior-mean} performs source-level borrowing by adaptively combining information from globally aligned source studies to construct a source-guided posterior for $\bm\beta$. The projection step via equation~\eqref{eq:projection} then performs coordinate-level refinement by inducing adaptive sparsity, preferentially retaining coefficients supported by strong agreement across multiple sources while shrinking source-specific or non-transferable signals. This two-stage design allows ProjectionTL to distinguish global source relevance from local variable-level transferability, thereby improving robustness against heterogeneous sources and mitigating negative transfer.

\subsection{Extension to Multi-Modality} 
\begin{figure}[htbp]
    \centering
    \includegraphics[width=0.7\linewidth]{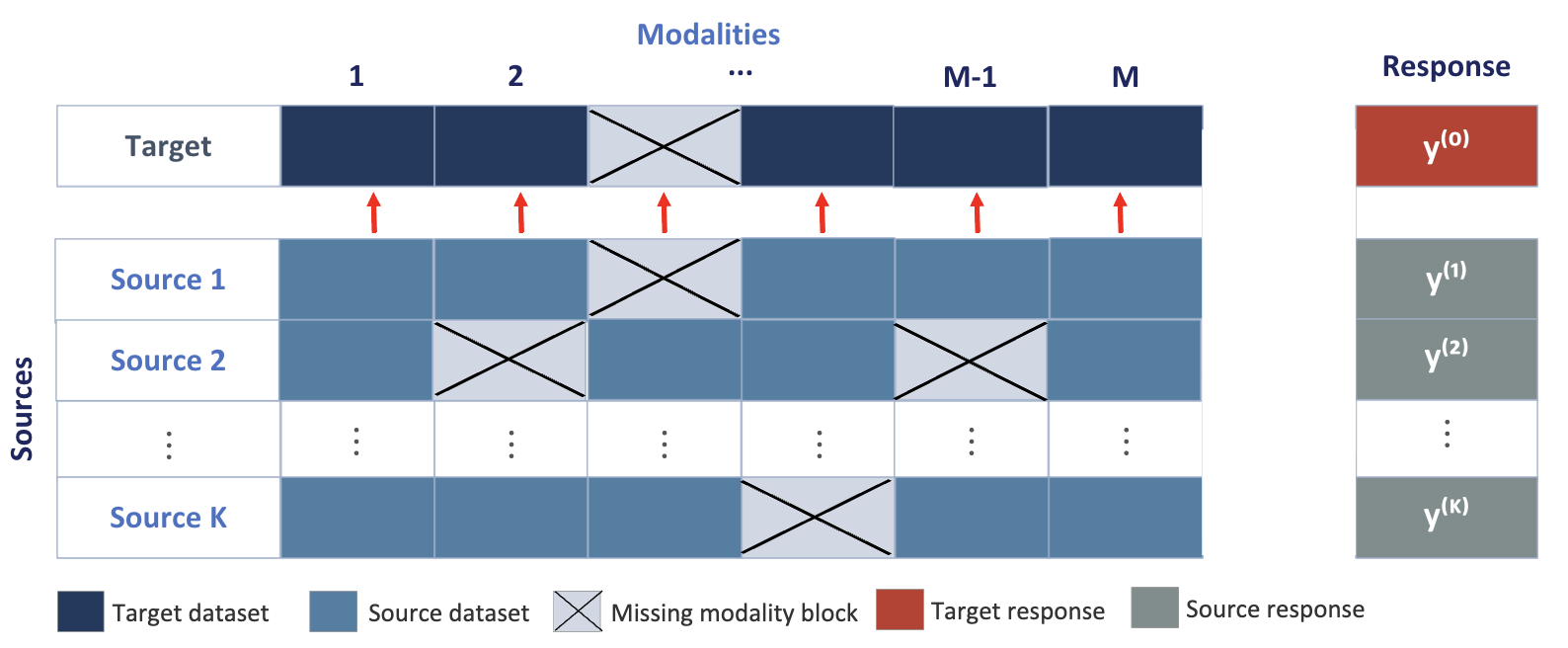}
    \caption{Illustration of multi-modal transfer learning with block-wise missing modalities.}
    \label{fig:multi-modal_diag}
\end{figure}

We extend ProjectionTL to multi-modal settings in which different sources observe different subsets of the modalities available in the target study, resulting in block-wise missingness (Figure~\ref{fig:multi-modal_diag}). Let the target covariates be partitioned into $M$ modality blocks, $\bm X=(\bm X^{(1)},\ldots,\bm X^{(M)})$, where modality $m$ contains $p_m$ covariates and $\sum_{m=1}^M p_m=p$. The corresponding target coefficient vector is partitioned as $\bm\beta=(\bm\beta^{(1)},\ldots,\bm\beta^{(M)})$. We assume that the target study observes all $M$ modalities, whereas each source may observe only a subset. For source $k$, let $\mathcal M_k\subseteq\{1,\ldots,M\}$ denote its observed modalities and $\mathcal I^{(m)}$ denote the covariate indices belonging to modality $m$. The set of covariates available in source $k$ is then $\mathcal I_k=\bigcup_{m\in\mathcal M_k}\mathcal I^{(m)}$.

ProjectionTL accommodates this structure by borrowing information only from the modalities observed in each source, without imputing missing modality blocks. However, because each source fits a reduced model based on its observed covariates, the resulting coefficient estimates may differ from the corresponding target full-model coefficients due to omitted-variable effects, even in the absence of genuine source-target heterogeneity. To address this issue, we introduce a source-specific calibration procedure that uses the fully observed target data to estimate the relationships between observed and omitted covariates, allowing the reduced-model source coefficients to be calibrated to the target full-model scale before applying ProjectionTL.

\subsubsection{Source Calibration Under Partially Overlapping Covariates}
\label{sec:source_calibration}

We calibrate each reduced-model coefficient onto the target full-model scale before applying projection.
For source $k$, write $\bm X_i=(\bm X_{i,\mathcal I_k}^{\mathrm T}, \bm X_{i,\mathcal I_k^c}^{\mathrm T})^{\mathrm T}$ and partition $\bm\beta^0=(\bm\beta_{\mathcal I_k}^{0\mathrm T}, \bm\beta_{\mathcal I_k^c}^{0\mathrm T})^{\mathrm T}$. The source provides only a reduced-model coefficient estimate
$\widehat{\bm\omega}_k\in\mathbb R^{|\mathcal I_k|}$. Following the reduced-model reparameterization of \citet{choi2024robust}, the target parameter comparable to the reduced-model coefficient is
$\bm\phi_k^0 = \bm\beta_{\mathcal I_k}^0 + \bm\Gamma_k \bm\beta_{\mathcal I_k^c}^0$, where $\bm\Gamma_k$ denotes the target-population projection of the omitted covariates onto the observed covariates. Detailed derivations and estimation of $\bm\Gamma_k$ are provided in Supplement~A.

Let $\widehat{\bm\beta}$ denote the target-only estimator obtained from the target data and let
$\widehat{\bm\phi}_k = \widehat{\bm\beta}_{\mathcal I_k} + \widehat{\bm\Gamma}_k \widehat{\bm\beta}_{\mathcal I_k^c}$.
Following the proportional-effect model of \citet{choi2024robust}, we estimate the source-specific scale
$\widehat c_k = {\widehat{\bm\omega}_k^{\mathrm T} \widehat{\bm\phi}_k}/{\|\widehat{\bm\omega}_k\|_2^2}$,
and define the calibrated source coefficient
$\widehat b_{kj} = \widehat c_k \widehat\omega_{kj} - (\widehat{\bm\Gamma}_k \widehat{\bm\beta}_{\mathcal I_k^c})_j$, with $j\in\mathcal I_k$. For unobserved coordinates $j\notin\mathcal I_k$, $\widehat b_{kj}$ is left undefined, so missing covariates are treated as unavailable information rather than observed zeros.
ProjectionTL is then applied to the calibrated coefficients
\(\widehat b_{kj}\). For source \(k\), Stage~1 computes its cosine
similarity using only its observed coordinates, $\theta_k
=({
\widehat{\bm b}_{k,\mathcal I_k}^{\mathrm T}
\widehat{\bm\beta}_{\mathcal I_k})/
}{
\|\widehat{\bm b}_{k,\mathcal I_k}\|_2
\|\widehat{\bm\beta}_{\mathcal I_k}\|_2
},$
with a small positive floor imposed when necessary to obtain a valid
Dirichlet concentration parameter. Let
\(\mathcal K_j=\{k:j\in\mathcal I_k\}\) and let \(\eta_0\) denote the
no-borrow weight. Conditional on the global source weights, define $d_j=\eta_0+\sum_{k\in\mathcal K_j}\eta_k,
\text{ and }
\mu_j(\bm\eta)
=
\sum_{k\in\mathcal K_j}
({\eta_k}/{d_j}) \cdot \widehat b_{kj}.$
Thus, a source that does not observe coordinate \(j\) is excluded from
both the numerator and the normalizing denominator rather than being
treated as having coefficient zero. Stage~2 similarly aggregates coordinate-wise agreement only across
observed source-coordinate pairs. Specifically, for
\(\mathcal K_j=\{k:j\in\mathcal I_k\}\neq\varnothing\), we define
$a_j=|\mathcal K_j|^{-1}
\sum_{k\in\mathcal K_j}
\exp\{
-(\widehat\beta_j-\widehat b_{kj})^2
\}
.$
Hence, sources that do not observe coordinate \(j\) do not contribute to
its projection weight, and differences in the number of sources observing
each coordinate do not mechanically alter its penalty. If
\(\mathcal K_j=\varnothing\), we set \(a_j=1\), so that coordinate \(j\)
receives the ordinary target-only projection penalty.
Consequently, ProjectionTL accommodates partially overlapping source
covariates without imputing the missing source modalities or interpreting
ordinary omitted-variable effects as source-target heterogeneity.

{ The calibrated coefficients $\widehat b_{kj}$ depend
on the target responses through $\widehat{\bm\beta}$, both in $\widehat c_k$ and in
$(\widehat\Gamma_k\widehat{\bm\beta}_{I_k^c})_j$, so calibration can change the direction, and not
only the length, of the source columns. To remove this dependence in the theory of Section~3, we
also consider a cross-fitted version: split the target rows into disjoint sets $\mathcal T_1$ and
$\mathcal T_2$ of sizes $n_{0,1}$ and $n_{0,2}=n_0-n_{0,1}$, with $n_{0,2}\asymp n_0$; compute
$\widehat{\bm\beta}$, $\widehat\Gamma_k$, $\widehat c_k$ and $\widehat b_{kj}$ from the rows in
$\mathcal T_1$ and the source estimates only; and run Stages~1 and~2 on
$(\bm Y_{\mathcal T_2},\bX_{\mathcal T_2})$. Conditionally on the rows in $\mathcal T_1$, the
calibrated coefficients are then fixed and independent of the errors in $\mathcal T_2$.}

\vspace{-1em}
\section{Theoretical Results}\label{sec:theory}
This section establishes posterior contraction, transfer-benefit conditions, model-selection consistency, and post-selection uncertainty validity for ProjectionTL. We use the equivalent $n_0$-normalized form of equation \eqref{eq:projection} to obtain $\bm\beta^\ast$, i.e., $(2n_0)^{-1}\|\bm X\bm\beta-\bm X\bm u\|_2^2+\lambda_n\sum_j |u_j|/a_j$, with $\lambda_n=\lambda/(2n_0)$ the correspondingly rescaled tuning parameter, as rescaling the objective by a positive constant does not change its minimizer $\bm\beta^\ast$.

\subsection{Estimation and Selection Consistency}

We first state the regularity conditions for posterior contraction and model selection, combining standard high-dimensional regression conditions on the target design with mild boundedness conditions on the source summaries and adaptive projection weights. Importantly, no source is required to consistently estimate the target parameter. 

\begin{assumption}[Target design and bounded variance]\label{design}\label{ass:design}
The target model satisfies
\(\bY=\bX\bm\beta^0+\bm\varepsilon\), where the errors \(\varepsilon_i\) are independent, mean zero, and sub-Gaussian with scale bounded by a constant \(\sigma_0\).  The columns of \(\bX\in\mathbb R^{n_0\times p}\) are normalized as
\(\|\bX^{(j)}\|_2=\sqrt{n_0}\), for \(j=1,\ldots,p\).  The variance value used in conditional posterior calculations satisfies \(\sigma^2\in\mathcal U_n=[\underline\sigma^2,\overline\sigma^2]\), where \(0<\underline\sigma^2<\overline\sigma^2<\infty\).  The source-specific drift induced by the prior precision \(\tau_n\) is controlled by Assumption~\ref{ass:source-bound}.
\end{assumption}
\begin{assumption}[Compatibility and cone restricted eigenvalue]\label{ass:compatibility}
Let \(d_j=1/\widetilde a_j\) and let \(d_{\min},d_{\max}\) be the deterministic bounds in Assumption~\ref{ass:weights-main}. Define
$C_D=({4d_{\max}+d_{\min}})/{3d_{\min}}.$
There exist constants \(\kappa>0\) and \(\kappa_2>0\) such that, for every \(\bm h\in\mathbb R^p\) satisfying
\(\|\bm h_{S_0^c}\|_1\le C_D\|\bm h_{S_0}\|_1\),
${n_0^{-1}}\|\bX\bm h\|_2^2
  \ge
  \kappa^2 {\|\bm h_{S_0}\|_1^2}/{s_0}\text{ and }
{n_0^{-1}}\|\bX\bm h\|_2^2
  \ge
  \kappa_2^2\|\bm h\|_2^2.$
\end{assumption}

\begin{assumption}[Bounded source summaries]\label{ass:source-bound}
For a deterministic sequence \(B_n\), with \(\mathbb P_0\)-probability tending to one, $ \max_{1\le k\le K}\|\widehat{\bm\omega}_k\|_2\le B_n
  \text{ and }
  n_0^{-1}{\tau_n(B_n+\|\bm\beta^0\|_2)}=o(\lambda_n).$
\end{assumption}
\begin{assumption}[Bounded adaptive projection weights]\label{ass:weights-main}\label{ass:weight-transfer}
The component-level projection uses the truncated weights
$\widetilde a_j=\min\{\overline a,\max(a_j,\underline a)\}, 0<\underline a<\overline a<\infty$. Equivalently, the penalty factors \(d_j=1/\widetilde a_j\) satisfy \(0<d_{\min}\le d_j\le d_{\max}<\infty\).
\end{assumption}
The contraction argument proceeds in two stages: the dense Stage 1 posterior is controlled through its target score, and the weighted projection in equation \eqref{eq:projection} translates this control into prediction and coefficient contraction under the compatibility conditions.



\begin{theorem}[Posterior contraction of the projected transfer posterior]\label{thm:main_contraction}
Suppose Assumptions~\ref{design}-\ref{ass:weights-main} hold, and let
\(\lambda_n=A\sqrt{(\log p)/n_0}\) for a sufficiently large constant \(A>0\).  Then there exists a constant \(M>0\) such that
{\small\[
\Pi_{\mathrm{ProjectionTL}}
\left(
  \frac{1}{n_0}\|\bX(\bm\beta^\ast-\bm\beta^0)\|_2^2
  >
  M\frac{s_0\log p}{n_0}
  \,\middle|\,
  \mathcal D_0,\mathcal D_1,\ldots,\mathcal D_K
\right)
\to 0
\]}
in \(\mathbb P_0\)-probability.  Moreover, under the cone restricted-eigenvalue part of Assumption~\ref{ass:compatibility},
{\small\[
\Pi_{\mathrm{ProjectionTL}}
\left(
  \|\bm\beta^\ast-\bm\beta^0\|_2
  >
  M\sqrt{\frac{s_0\log p}{n_0}}
  \,\middle|\,
  \mathcal D_0,\mathcal D_1,\ldots,\mathcal D_K
\right)
\to 0.
\]}
\end{theorem}
Theorem~\ref{thm:main_contraction} establishes a safe-transfer guarantee: ProjectionTL retains the standard target-only sparse-regression rate even when sources are heterogeneous or uninformative, without assuming oracle source weights. Without additional assumptions on source-target similarity, a uniformly faster rate cannot generally be guaranteed, since the target-only problem is included as a special case.{Theorem~\ref{thm:main_contraction} does not require the learned Dirichlet weights to concentrate: its guarantee holds uniformly over the transfer weights, which is what protects against negative transfer. Concentration of the weights around the pseudo-true transfer target (Lemma~F.6) is used only for the source-dependent improvement in Corollary~\ref{cor:beneficial-transfer}.} When stronger source-closeness conditions hold, Corollary~\ref{cor:beneficial-transfer} below further quantifies the benefit of informative sources through a sharper source-dependent bound{, under stronger source anchoring ($\tau_n/p\to\infty$) than in Theorem~\ref{thm:main_contraction}}. This distinction complements existing transfer-learning guarantees. For example, sharper rates in Trans-LASSO \citep{li2022transfer} and TRADER \citep{lai2026bayesian} rely on favorable source-target relationships, such as an informative-source set or an oracle source combination. ProjectionTL {obtains} robustness without such favorable-source assumptions{, through uniform control of the source drift rather than a separate source-detection step,} and then characterizes the additional gain available when informative sources are present. Detailed comparisons and technical discussion are deferred to the Supplement.
Let \(\bW=[\widehat{\bm\omega}_1,\ldots,\widehat{\bm\omega}_K,\bm 0]\) denote the rescaled source dictionary from Section~\ref{sec:method}.  Conditional on \((\bm\eta,\sigma^2)\), the Stage 1 posterior satisfies $\bm\beta\mid \bY,\bm\eta,\sigma^2
  \sim
  N_p\{\bm m_\eta,\sigma^2\bV_n\}
  \text{ with }
  \bV_n=(\bX^\mathrm T\bX+\tau_n\bm I_p)^{-1}$
and \(\bm m_\eta=\bV_n(\bX^\mathrm T\bY+\tau_n\bW\bm\eta)\). The posterior density of \(\bm\eta\) is proportional to $(\prod_{k=1}^{K+1}\eta_k^{\theta_k-1})
  \exp\{-\frac12(\bY-\bX\bW\bm\eta)^\mathrm T
  \bm\Sigma^{-1}(\bY-\bX\bW\bm\eta)\}
  \text{ with }
  \bm\Sigma=\sigma^2(\bm I_{n_0}+\tau_n^{-1}\bX\bX^\mathrm T).$
\begin{assumption}[Identifiability of transfer weights]\label{eta_parameter_space}\label{identifiability}\label{margin}
The transfer weights satisfy \(\bm\eta\in\Delta^{K+1}\), and the prior on \(\bm\eta\) assigns positive mass to every relative open neighborhood of every point in \(\Delta^{K+1}\). Let $\bm A_n=\left(\bm I_{n_0}+\tau_n^{-1}\bX\bX^\mathrm T\right)^{-1} \text{ and }
\widetilde R_n(\bm\eta)
={n_0^{-1}}\{\bX(\bm\beta^0-\bW\bm\eta)\}^\mathrm T
\bm A_n
\{\bX(\bm\beta^0-\bW\bm\eta)\}.$
There exists a unique minimizer
\(\bm\eta^\star=\arg\min_{\bm\eta\in\Delta^{K+1}}\widetilde R_n(\bm\eta)\).  In addition, for every \(\epsilon>0\), there is a constant \(c_\epsilon>0\) such that, with probability tending to one, $\inf_{\|\bm\eta-\bm\eta^\star\|_2\ge \epsilon}
\{\widetilde R_n(\bm\eta)-\widetilde R_n(\bm\eta^\star)\}
\ge c_\epsilon.$
\end{assumption}
\begin{assumption}[Row-space spectrum of the target design]\label{ass:row-space}
Let $p\ge n_0$. There exist constants $0<c_X\le C_X<\infty$ such that, with
$P_0$-probability tending to one, $  c_X\,({p}/{n_0})
  \;\le\; \lambda^{+}_{\min}\!\Big({n_0^{-1}}\bm X\bm X^{\mathrm T}\Big)
  \;\le\; \lambda_{\max}\!\Big({n_0^{-1}}\bm X\bm X^{\mathrm T}\Big)
  \;\le\; C_X\,({p}/{n_0}),$
equivalently, every nonzero eigenvalue of $\bm X\bm X^{\mathrm T}$ lies in
$[c_X\,p,\;C_X\,p]$.
\end{assumption}
Assumption~\ref{margin} ensures that the Stage~1 likelihood identifies a unique best source/no-source mixture. Importantly, \(\bm\eta^\star\) may lie anywhere on the simplex: it includes full transfer, no transfer, and partial transfer \(0<\eta_{K+1}^\star<1\). Assumption~\ref{ass:row-space} provides standard spectral control of the target design needed to quantify the effect of source anchoring.

To characterize when transfer is beneficial, define the prediction discrepancy at the pseudo-true mixture $\rho_n^\star = {n_0}^{-1} \|\bX(\bm\beta^0-\bW\bm\eta^\star)\|_2^2$. Thus, \(\rho_n^\star\) directly measures how well the {pseudo-true} source mixture {targeted} in Stage~1 approximates the target signal in prediction space. Because the source columns of $\bW$ are rescaled to the length of the target pilot, $\rho_n^\star$ also reflects how accurately that length is estimated; a strict improvement therefore requires both close sources and an accurate pilot scale. Define $q_n=\sqrt{({n_0\rho_n^\star})/{\log p}}$, $ \gamma_n(\tau_n)={C_X\,p}/({C_X\,p+\tau_n})$, $\alpha_n(\tau_n)={\tau_n}/({c_X\,p+\tau_n})$
and $\mathcal M_n(\tau_n,q_n)=\sqrt{\gamma_n(\tau_n)}+\alpha_n(\tau_n)\,q_n$.  The first term {is the target-sample noise that remains after source anchoring}, whereas the second captures the cost of source-target mismatch. Hence, transfer is beneficial when the reduction in noise outweighs the discrepancy introduced by the source-centered prior. Unlike TRADER \citep{lai2026bayesian}, which obtains faster contraction rates when the target coefficient differs from a source-centered coefficient vector on only a sparse set of coordinates, ProjectionTL characterizes transferability through the prediction discrepancy of the learned source mixture. Specifically, the coefficient discrepancy $\bm\beta^0-\bW\bm\eta^\star$ may be dense, provided its induced prediction error $\rho_n^\star$ is sufficiently small. Consequently, the transfer benefit in ProjectionTL is reflected in the source-anchoring multiplier $\mathcal M_n(\tau_n,q_n)$, rather than a reduction in the effective sparsity level $s_0$. This distinction is particularly relevant for correlated high-dimensional designs, where substantial coefficient differences may correspond to small differences in predicted outcomes. The two approaches therefore characterize different, non-nested notions of transferability. Under sufficiently strong source anchoring and control of the source-weight estimation error, $\rho_n^\star=o\{(\log p)/n_0\}$ yields a strict order improvement over the target-only contraction rate.

We next characterize when informative sources can improve upon the safe contraction rate in Theorem~\ref{thm:main_contraction}. The key trade-off is between two effects of source anchoring: it reduces the contribution of target-sample noise, but introduces error when the learned source mixture differs from the target. The resulting transfer factor $\mathcal M_n(\tau_n,q_n)$ balances these two effects, yielding a sharper contraction bound when the pseudo-true source mixture is sufficiently close to the target {and the source anchoring is strong}.

\begin{corollary}[Beneficial transfer through source anchoring]\label{cor:beneficial-transfer}
Suppose Assumptions~\ref{design}, \ref{ass:compatibility}, \ref{ass:weights-main}-\ref{ass:row-space} {and Assumption~10 in the Supplement} hold and $G_n^2 := \sup_{\bm\eta\in\Delta^{K+1}} {n_0^{-1}} \|\bX(\bm\beta^0-\bW\bm\eta)\|_2^2 = O(1)$. 
{Let $M_n\to\infty$ be any sequence, and let the projection be tuned at}
{$\lambda_n''=A\Big\{\sqrt{\gamma_n(\tau_n)}\sqrt{{{n_0^{-1}}\log p}}+\alpha_n(\tau_n)\sqrt{\rho_n^\star}+\sqrt{{{n_0^{-1}}\alpha_n(\tau_n)M_n}}\Big\}$}
{for a sufficiently large constant $A>0$.}
Then there exists a constant $M>0$ such that, in $P_0$-probability,
{\small\[
\Pi_{\mathrm{ProjectionTL}}
\left(
\|\bm\beta^\ast-\bm\beta^0\|_2
>
M\sqrt{\frac{s_0\log p}{n_0}}\,
{\Big\{}\mathcal M_n(\tau_n,q_n){+\sqrt{\frac{\alpha_n(\tau_n)M_n}{\log p}}\Big\}}
\,\middle|\,
\mathcal D_0,\mathcal D_1,\ldots,\mathcal D_K
\right)
\to0{.}
\]}
{If $M_n=o(\log p)$, the second term in the braces is $o(1)$.}
\end{corollary}
Corollary~\ref{cor:beneficial-transfer} complements the safe-transfer result of
Theorem~\ref{thm:main_contraction}. Theorem~\ref{thm:main_contraction} guarantees
the target-only sparse rate without requiring informative sources, while the
corollary shows how the contraction bound can improve when the learned source
mixture is close to the target. The two results describe different regimes of the
prior precision: Assumption~{\ref{ass:source-bound}} keeps $\tau_n$ small enough that the source-centered
prior cannot harm the target rate, whereas a gain requires stronger source
anchoring. Because the nonzero eigenvalues of {$\bX\bX^\mathrm T$} are of
order $p$, anchoring reduces the target-noise contribution only once $\tau_n$ is
large relative to $p$, not merely to $n_0$. In the source-dominant regime
$\tau_n/p\to\infty$, $\gamma_n(\tau_n)\to0$ and $\alpha_n(\tau_n)\to1$, and the
bound becomes
{$\|\bm\beta^\ast-\bm\beta^0\|_2
=O_\Pi\big(\sqrt{s_0\rho_n^\star}+\sqrt{s_0\gamma_n(\tau_n)(\log p)/n_0}+\sqrt{s_0M_n/n_0}\big)$}.
Thus, if $\rho_n^\star=o\{(\log p)/n_0\}$ {and $M_n=o(\log p)$}, the source-anchored bound is of strictly
smaller order than the target-only bound, i.e.\ $\mathcal M_n(\tau_n,q_n)\to0$;
since both rates hold up to constants, it is this order improvement, rather than
$\mathcal M_n<1$ alone, that quantifies the benefit of informative sources.
{The tuning parameter $\lambda_n''$ depends on the unknown $\rho_n^\star$, so Corollary~\ref{cor:beneficial-transfer} describes what the procedure can attain under suitable tuning rather than a fully data-driven guarantee.}
The notion of transferability here differs from residual-sparsity conditions used
in methods such as TRADER. ProjectionTL requires the learned source mixture to be
close to the target in prediction space, while the coefficient discrepancy
$\bm\beta^0-\bW\bm\eta^\star$ itself may remain dense. This prediction-level notion
of alignment is particularly natural under correlated high-dimensional designs.

For exact support recovery, we need stronger conditions than those required for contraction. As in weighted LASSO selection, we require that inactive variables are sufficiently separated from the active set, that the nonzero coefficients are large enough to be distinguishable from zero, and that the projection penalty is sufficiently strong. Let \(S_0=\{j:\beta_j^0\ne0\}\) denote the true support and \(\bC_n=n_0^{-1}\bX^\mathrm T\bX\). Partition \(\bC_n\) according to \((S_0,S_0^c)\) as \(\bC_n=(C_{n(ab)})_{a,b\in\{1,2\}}\), and let \(D=\operatorname{diag}(d_1,\ldots,d_p)\), where \(d_j=1/\widetilde a_j\).

\begin{assumption}[Weighted irrepresentability]\label{ass:selection-design}
There exists \(\xi\in(0,1)\) such that, with probability tending to one, $\|
  D_{S_0^c}^{-1}
  C_{n(21)}C_{n(11)}^{-1}
  D_{S_0}\operatorname{sgn}(\bm\beta^0_{S_0})
\|_\infty
\le 1-\xi
\text{ and }
\lambda_{\min}(C_{n(11)})\ge c_{\min}>0,
\|C_{n(11)}^{-1}\|_\infty\le C_{\mathrm{inv}}<\infty,\text{ and }
\|D_{S_0^c}^{-1}C_{n(21)}C_{n(11)}^{-1}\|_\infty\le C_{\mathrm{sel}}<\infty.$
\end{assumption}

\begin{assumption}[Selection tuning and beta-min]\label{ass:tuning-transfer}
The selection-stage tuning parameter satisfies $\lambda_n=M_n\sqrt{{(\log p)}/{n_0}},
  M_n\to\infty,
  \lambda_n\to0.$
In addition, $\beta_{\min}:=\min_{j\in S_0}|\beta^0_j|
  \gg \lambda_n.$
\end{assumption}

\begin{assumption}[Projection uniqueness]\label{ass:projection-unique}
The weighted projection solution is unique with probability tending to one.  Otherwise, \(\bm\beta^\ast\) denotes the minimizer selected by a fixed deterministic tie-breaking rule.
\end{assumption}
\noindent Assumption~\ref{ass:selection-design} is the weighted-LASSO analogue of the standard irrepresentability condition and prevents inactive variables from being too strongly represented by the active set after accounting for the adaptive penalties. Assumption~\ref{ass:tuning-transfer} strengthens the projection penalty for exact selection and requires the true signals to remain distinguishable from zero. Assumption~\ref{ass:projection-unique} ensures that the selected support is well defined. 
Let \(\widehat S^\ast=\{j:\beta^\ast_j\ne0\}\).
\begin{theorem}[Model selection consistency]\label{thm:ms-consis-transfer}
Under Assumptions~\ref{design}, \ref{ass:source-bound}, \ref{ass:weights-main}, and \ref{ass:selection-design}-\ref{ass:projection-unique}, we have $\inf_{\sigma^\ast\in\mathcal U_n}
  \Pi\bigl(\widehat S^\ast=S_0\mid \bY,\sigma^\ast\bigr)
  \to 1\text{ in }P_{\bm\beta^0}\text{-probability}.$
\end{theorem}

\noindent Theorem~\ref{thm:ms-consis-transfer} shows that the projected posterior consistently recovers the true support even in the presence of heterogeneous or uninformative sources. Importantly, the result does not require known informative source sets or assume accurate source estimates, while source heterogeneity is accommodated through the source-weighting and projection mechanisms, and support recovery is guaranteed under standard weighted high-dimensional selection conditions. The adaptive projection weights can provide an additional selection advantage when the sources preferentially support true signals: larger \(a_j\) on active coordinates reduces their effective penalties relative to noise variables. Corollary 3.1 in Supplement F
formalizes conditions under which this source information relaxes the corresponding weighted irrepresentability and signal-strength requirements relative to a source-blind LASSO. 

\subsection{Coverage of Post-selection Credible Ellipsoid}
Uncertainty quantification is nontrivial here because the Stage 1 posterior is centered at a data-driven, source-informed prior mean.  If that source-centered shift persisted at the post-selection scale, it could distort frequentist calibration.  The next result shows that, after support recovery, the source-prior shift is asymptotically negligible in the selected coordinates, so the usual post-selection ellipsoid has the same limiting calibration as in the target-only sparse projection posterior \citep{pal2024bayesian}.
{Let $\widehat S=\{j:\Pi(\beta^\ast_j\neq0\mid\bY)>1/2\}$ denote the median
probability model support of Section~2.3, and $\widehat Q:=\bX_{\widehat S}$. The reported
credible set is built on $\widehat S$: let $\widehat{\bm\beta}^{\mathrm{R,PS}}
=(\widehat Q^{\mathrm T}\widehat Q+\tau_n\bm I_{|\widehat S|})^{-1}\widehat Q^{\mathrm T}\bY$, and let
$\bm\beta^{\mathrm{PS}}_{\widehat S}$ be the restricted Stage-1 Gaussian posterior draw on $\widehat S$,
which, conditional on $(\bm\eta,\sigma^2,\bY)$, has mean
$(\widehat Q^{\mathrm T}\widehat Q+\tau_n\bm I_{|\widehat S|})^{-1}(\widehat Q^{\mathrm T}\bY+\tau_n\bW_{\widehat S}\bm\eta)$
and covariance $\sigma^2(\widehat Q^{\mathrm T}\widehat Q+\tau_n\bm I_{|\widehat S|})^{-1}$. This is a
separate draw, not the non-Gaussian projected draw $\bm\beta^\ast$. For the analysis we also use the
corresponding oracle quantities on the true support} $S_0=\{j:\beta^0_j\neq0\}$, $|S_0|=s_0$: define
$Q:=\bX_{S_0}\in\mathbb R^{n_0\times s_0}${, and let $\widehat{\bm\beta}^{\mathrm{R,PS}}$,
$\bm\beta^{\mathrm{PS}}_{S_0}$ denote the same objects with $\widehat S$ replaced by $S_0$.} The
restricted posterior on $S_0$ has mean
$\mu^{\mathrm{PS}}_{\eta}=(Q^{\mathrm T}Q+\tau_n\bm I_{s_0})^{-1}(Q^{\mathrm T}\bY+\tau_n\bW_{S_0}\bm\eta)
=\widehat{\bm\beta}^{\mathrm{R,PS}}+\delta_\eta$, where $\delta_\eta:=\tau_n(Q^{\mathrm T}Q+\tau_n\bm I_{s_0})^{-1}\bW_{S_0}\bm\eta$
captures the additional pull of the source-centered prior on the selected-model posterior mean,
and covariance $\sigma^2(Q^{\mathrm T}Q+\tau_n\bm I_{s_0})^{-1}$.

\begin{lemma}[Negligibility of the source-prior shift and the ridge bias]\label{lem:delta-negligible}
Assume \(\lambda_{\min}(\Q^\mathrm T\Q/n_0)\ge c_0>0\) and Assumption~\ref{ass:source-bound}.  Conditional on \(\widehat S^\ast=S_0\), whenever \(\tau_n(B_n+\|\bm\beta^0\|_2)=o(\sqrt{n_0})\),
we have $\E_{\boldeta\mid\bY}\|\deltaeta\|_2
=o_p(n_0^{-1/2})$ and $\big\|\tau_n(\Q^\mathrm T\Q+\tau_n\bm I_{s_0})^{-1}\bm\beta^0_{S_0}\big\|_2
=o(n_0^{-1/2}).$
In particular, $\E_{\boldeta\mid\bY}
\left[
\deltaeta^\mathrm T(\Q^\mathrm T\Q+\tau_n\bm I_{s_0})\deltaeta
\right]
=o_p(1).$
\end{lemma}
\noindent According to Lemma~\ref{lem:delta-negligible}, the restricted posterior has the same first-order centering as the target-only Gaussian approximation. Lemma~\ref{lem:delta-negligible} shows that the source-induced posterior shift and ridge bias are
asymptotically negligible, while Lemma~\ref{lem:r-alpha} establishes convergence of the posterior
quadratic-form quantile to the corresponding chi-square quantile. 
\begin{lemma}[Convergence of the posterior quantile]\label{lem:r-alpha}
Assume the conditions of Lemma~\ref{lem:delta-negligible}, \(s_0\) fixed, \(\Q^\mathrm T\Q/n_0\to C_{S_0}\) with \(C_{S_0}\) positive definite, and posterior consistency of \(\sigma^2\).  If \(r_\alpha\) denotes the posterior \((1-\alpha)\)-quantile of $T=(\bm\beta_{S_0}^{\mathrm{PS}}-\widehat{\bm\beta}^{\mathrm{R,PS}})^\mathrm T
(\Q^\mathrm T\Q+\tau_n\bm I_{s_0})
(\bm\beta_{S_0}^{\mathrm{PS}}-\widehat{\bm\beta}^{\mathrm{R,PS}}),$
then $r_\alpha\to \sigma_0^2\chi^2_{s_0,1-\alpha} \text{ in }P_0\text{-probability}.$
\end{lemma}
Combining the negligibility of the source-induced centering error with the convergence of the posterior quadratic-form quantile yields the following post-selection coverage guarantee. It should be noted that the coverage result invokes support recovery, not sign consistency. 
\begin{theorem}[Asymptotically correct coverage of the credible ellipsoid]\label{thm:coverage}
Assume the support-recovery conditions of Theorem~\ref{thm:ms-consis-transfer}, the conditions of
Lemmas~3.1-3.2, {and that $\sigma_0^2$ lies in the interior of $\mathcal U_n$.} In
addition, for the coverage calculation assume $\varepsilon_i\overset{iid}{\sim}N(0,\sigma_0^2)$.
{Let $\widehat r_\alpha$ denote the posterior $(1-\alpha)$-quantile of
$\widehat T=(\bm\beta^{\mathrm{PS}}_{\widehat S}-\widehat{\bm\beta}^{\mathrm{R,PS}}_{\widehat S})^{\mathrm T}
(\widehat Q^{\mathrm T}\widehat Q+\tau_n\bm I_{|\widehat S|})
(\bm\beta^{\mathrm{PS}}_{\widehat S}-\widehat{\bm\beta}^{\mathrm{R,PS}}_{\widehat S})$.} Then the
post-selection credible ellipsoid ${\widehat D^{\mathrm{PS}}_{n,1-\alpha}=\Big\{\bm b\in\mathbb R^{|\widehat S|}:
(\bm b-\widehat{\bm\beta}^{\mathrm{R,PS}}_{\widehat S})^{\mathrm T}(\widehat Q^{\mathrm T}\widehat Q+\tau_n\bm I_{|\widehat S|})
(\bm b-\widehat{\bm\beta}^{\mathrm{R,PS}}_{\widehat S})\le\widehat r_\alpha\Big\}}$
satisfies ${P_0\big(\widehat S=S_0\ \text{and}\ \bm\beta^0_{\widehat S}\in\widehat D^{\mathrm{PS}}_{n,1-\alpha}\big)\to1-\alpha .}$
\end{theorem}
Theorem~\ref{thm:coverage} establishes asymptotically correct frequentist coverage for credible ellipsoids constructed from the restricted Gaussian posterior after support recovery. Under the stated conditions, the influence of the source-centered prior becomes asymptotically negligible, preserving nominal coverage despite adaptive information borrowing. Together with the preceding contraction and model selection results, this provides theoretical support for ProjectionTL in terms of estimation accuracy, variable selection, and post-selection uncertainty quantification.

\vspace{-1em}
\section{Simulation Study}\label{sec:simu}
The simulation studies evaluate the ability of ProjectionTL to improve target
estimation by selectively borrowing from related sources while remaining robust
to negative transfer from heterogeneous or misleading sources. We focus on three
criteria: estimation accuracy, support recovery, and uncertainty quantification.
Estimation accuracy is measured by mean squared error (MSE) between the
estimated and true target coefficient vectors, and support recovery is evaluated
using the Matthews correlation coefficient (MCC), which provides a balanced
summary of true and false discoveries. Supplementary analyses further examine performance in a fixed-feature-overlap
setting with both informative and non-informative sources
(Supplement~B), the impact of using the full target data
to construct the target pilot rather than a validation split
(Supplement~C), and sensitivity to the projection
tuning parameter \(\lambda\) (Supplement~D).
\subsection{Simulation Study 1: Overlap by Proportion of Signals}\label{subsec:simu_setup1}
In this setup, the degree of overlap is controlled by an overlap ratio parameter $\omega\in[0,1]$, representing the fraction of the target support shared with each source. Informative sources therefore contribute a varying proportion of useful predictors, with the remainder being filled with spurious non-overlap signals. This design emphasizes relative overlap proportion as the driver of transferability, while also introducing systematic spurious features. We consider a target regression task and $K$ source tasks with a shared predictor dimension $p$. The target response follows
$
Y_0 = X_0 \bm\beta^0 + \varepsilon_0,\;
\varepsilon_0 \sim \mathcal N_{n_0}(0,\sigma^2 I_{n_0}),
$
where $X_0 \in \mathbb R^{n_0\times p}$. The target coefficient vector $\bm\beta^0\in\mathbb R^p$ is $s$-sparse with support $S=\{1,\dots,s\}$; we set $\bm\beta^0=(\beta^0_1,\ldots,\beta^0_p)^\mathrm{T},\text{ such that }
\beta^0_j= 0.5\,S_j, \text{ if } j\in S, \textnormal{ and }
0, \text{ if } j\notin S$
where the target support $S=\{1,\ldots,s_0\}$ with $s_0=20$, and $S_j$ are i.i.d. Rademacher signs $\mathbb{P}(S_j=\pm1)=1/2$. Each source $k=1,\dots,K$ is generated as
$
Y_k = X_k\,\omega^{(k)} + \varepsilon_k,\;
\varepsilon_k \sim \mathcal N_{n_k}(0,\sigma^2 I_{n_k}),
$
with $X_k\in\mathbb R^{n_k\times p}$. Target and source covariates are drawn from $\mathcal N_p(0, \Sigma)$ with an AR(1) correlation structure, $\Sigma_{jk} = \rho^{|j-k|}$ with $\rho = 0.9$, so that neighbouring predictors are strongly collinear. We create partial transfer by letting each source overlap with the target on a subset $O_k\subset S$ of size $|O_k|=\lfloor \omega s_0\rfloor$ and carry source-specific spurious signals on a disjoint set $N_k\subset S^c$ with $|N_k|=n_{\text{spur}}$. The source coefficients are \(\omega^{(k)}_j=\beta^0_j\mathbf{1}_{\{j\in O_k\}}+2\zeta^{(k)}_j\mathbf{1}_{\{j\in N_k\}} \textnormal{ with } \zeta^{(k)}_j\overset{\mathrm{i.i.d.}}{\sim}\operatorname{Unif}\{-1,+1\}.\)
Thus, the target-overlapping effects are identical copies of $\beta^0$ on $O_k$, while $N_k$ injects non-transferable signals outside $S$. Across sources, $O_k$ and $N_k$ are sampled uniformly without replacement from $S$ and $S^c$, respectively. We vary the overlap ratio $\omega\in\{0.6,0.7,0.8,0.9,1.0\}$, where $|O_k|=\lfloor \omega s_0\rfloor$ for every source $k$. Unless stated otherwise, the remaining parameters are fixed at $
n_0=100,\; n_k=100,\; p=200,\; K=10,\; s_0=20,\;
n_{\text{spur}}=20,\; \sigma=1,\; b_{\text{mag}}=2.5.
$
For each $\omega$, we generate $100$ independent datasets (replicates). All LASSO fits use 10-fold cross-validation and default glmnet settings.


We compare four transfer strategies: ProjectionTL, TRADER, TransGLM and CONCERT, which is the variational spike-and-slab partial transfer model. Across Monte Carlo replications, we record the estimation error by $\mathrm{MSE}(\widehat\beta)=\|\widehat\beta-\beta^0\|_2^2/p$, support recovery by MCC between the estimated support and $S$ (for methods without inherent selection such as TRADER and CONCERT, we apply the natural threshold used in the code) and posterior interval coverage and length (95\%) for Bayesian methods (ProjectionTL, TRADER, CONCERT). We also check the source selection diagnostic: number of sources a method effectively ``borrows" from (e.g., TransGLM’s selected sources; ProjectionTL/TRADER via $\eta$ weights; CONCERT via average transfer probability). 

Varying $\omega\in[0.6,1]$ controls the degree of partial transferability while keeping effect sizes fixed; $\omega=1$ recovers full overlap, whereas smaller $\omega$ stresses negative-transfer resistance. The spurious set size $n_{\text{spur}}=20$ injects meaningful but moderate source-only signals to test selectivity. $p=200, s_0=20$ yields a sparse, moderately high-dimensional regime with $n_0\approx p$, where transfer can help but over-borrowing hurts. Equal signal magnitudes $|\beta^0_j|=0.5$ fix SNR, so differences arise from transfer behavior rather than target difficulty. Common noise level $\sigma=1$ isolates method differences from heteroskedastic confounders.

\begin{figure}[htbp]
    \centering
    \includegraphics[width=0.9\linewidth]{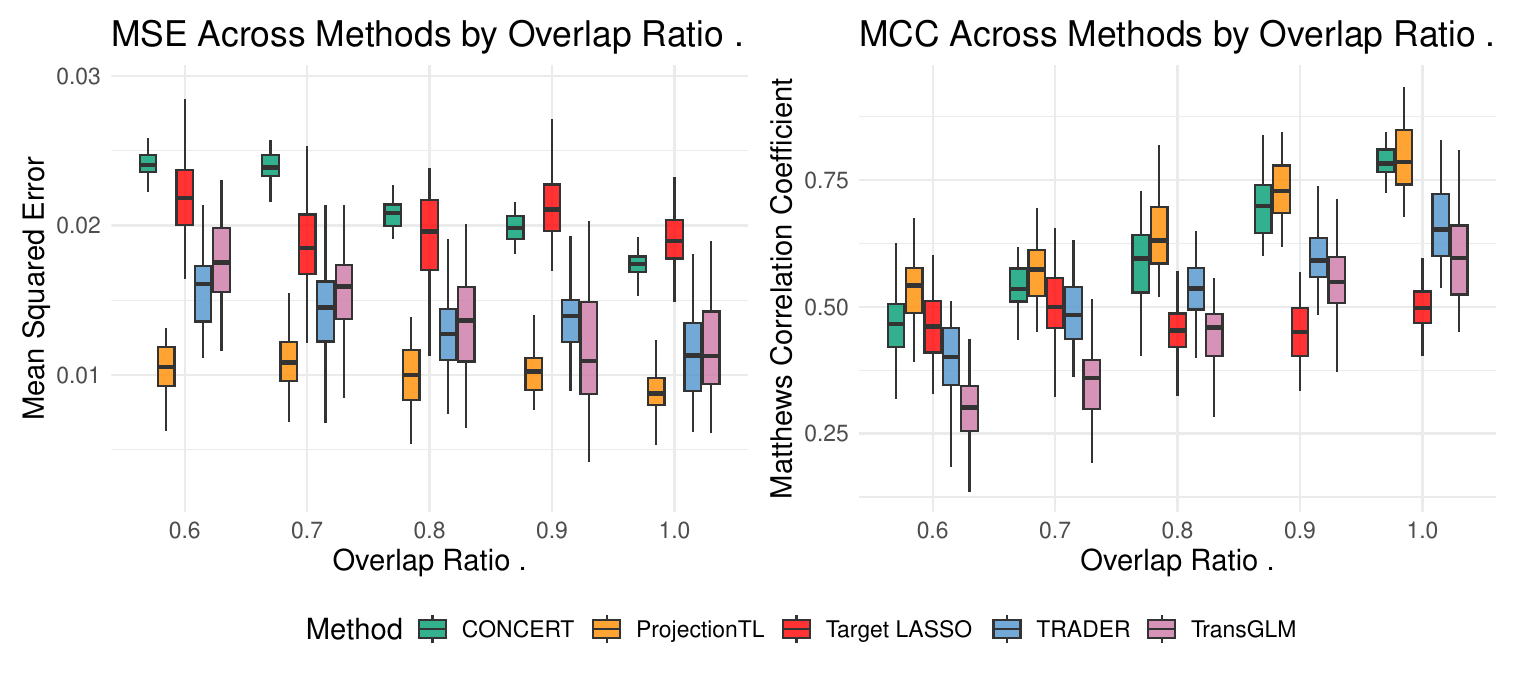}
    \caption{Average estimation error (MSE) and support recovery (MCC) as functions of the overlap ratio $\omega \in [0.6,1]$. Results are averaged over 100 Monte Carlo replicates with $n_0 = n_k = 100, p = 200, s_0 = 20$.}
    \label{fig:overlap_ratio}
\end{figure}
\begin{table}[ht]
\centering
\caption{Empirical 95\% interval coverage in Simulation~1 by overlap ratio $\omega$:
marginal coverage averaged over coordinates (MC), joint coverage of all $s_0$ true signals (JC),
and mean interval width (W).}
\label{tab:sim1-coverage}
\resizebox{\textwidth}{!}{%
\begin{tabular}{l *{5}{ccc}}
\toprule
& \multicolumn{3}{c}{$\omega=0.6$} & \multicolumn{3}{c}{$\omega=0.7$} & \multicolumn{3}{c}{$\omega=0.8$}
& \multicolumn{3}{c}{$\omega=0.9$} & \multicolumn{3}{c}{$\omega=1.0$} \\
\cmidrule(lr){2-4}\cmidrule(lr){5-7}\cmidrule(lr){8-10}\cmidrule(lr){11-13}\cmidrule(lr){14-16}
Method & MC & JC & W & MC & JC & W & MC & JC & W & MC & JC & W & MC & JC & W \\
\midrule
CONCERT      & .64 & .06 & .814 & .68 & .16 & .868 & .68 & .08 & .869 & .66 & .12 & .865 & .68 & .16 & .846 \\
TRADER       & .84 & .44 & .511 & .89 & .66 & .509 & .88 & .54 & .528 & .86 & .50 & .519 & .84 & .44 & .513 \\
TransGLM     & .82 & .50 & .393 & .83 & .56 & .395 & .81 & .45 & .407 & .82 & .42 & .407 & .84 & .42 & .399 \\
\cellcolor{gray!15}\textbf{ProjectionTL} & \cellcolor{gray!15}.94 & \cellcolor{gray!15}.80 & \cellcolor{gray!15}.680 & \cellcolor{gray!15}.94 & \cellcolor{gray!15}.82 & \cellcolor{gray!15}.684 & \cellcolor{gray!15}.97 & \cellcolor{gray!15}.83 & \cellcolor{gray!15}.739 & \cellcolor{gray!15}.93 & \cellcolor{gray!15}.67 & \cellcolor{gray!15}.735 & \cellcolor{gray!15}.86 & \cellcolor{gray!15}.71 & \cellcolor{gray!15}.601 \\
\bottomrule
\end{tabular}%
}
\end{table}
{As shown in Figure~\ref{fig:overlap_ratio}}, as the overlapping ratio $\omega$ increases, all methods benefit from greater signal overlap, but ProjectionTL consistently achieves the lowest MSE and highest MCC across all overlap levels. The performance gap is especially pronounced at moderate overlap ($\omega \in [0.7, 0.9]$), where ProjectionTL effectively exploits partial alignment while avoiding spurious borrowing. In contrast, methods such as TRADER and TransGLM exhibit higher estimation error and weaker support recovery when overlap is incomplete, reflecting their limited ability to selectively borrow only aligned components. CONCERT performs competitively in some regimes but remains less stable than ProjectionTL. Overall, these results demonstrate that ProjectionTL can robustly leverage partial overlap, achieving both improved estimation accuracy and superior variable selection compared to existing transfer-learning approaches, particularly when sources are only partially informative.

\subsection{Simulation Study 2: Multi-Modal Transfer Learning}
In this simulation study, we further evaluate the robustness under modality-level block missingness. We consider
\(
M=4 \ \text{modalities},\enskip
K=8 \ \text{sources},\enskip
p_m=40 \ \text{features per modality }(m=1,\dots,M).
\) Sample sizes are
\(
n_0 = n_k=200 \ \text{(per source)}. 
\) Thus the target design is $\bX_0 \in \R^{n_0 \times p}$, where $p = M\cdot p_m = 160.$ The source-level block-missingness parameter is \(L\in\{1,2,3\}\) (each source observes exactly \(L\) of the \(M\) modalities). Noise standard deviation varies as \(\sigma_\varepsilon\in\{0.5,1,2\}\). The number of shared active coordinates per modality is \(I\in\{0,1,2\}\). For each modality \(m\) we use a stationary AR(1) covariance with \(\rho=0.3\):
\(
\bm\Sigma_m \in\mathbb{R}^{p_m\times p_m},\enskip
(\Sigma_m)_{ij}=\rho^{\,|i-j|},\ \ i,j=1,\dots,p_m .
\) For each modality \(m\), we draw a 2-sparse coefficient vector \(\beta^{(m)}\in\mathbb{R}^{p_m}\) as follows:
pick an active set \(A_m\subset\{1,\dots,p_m\}\) with \(|A_m|=2\), and set $\beta^{(m)}_j=\sqrt{2}\,\mathbf{1}_{\{j\in A_m\}} \text{ so that } \|\bm\beta^{(m)}\|_2=2.$ Next, we generate the target design matrices
\(
\bX_0^{(m)}\in\mathbb{R}^{n_0\times p_m},\enskip
\text{where rows are i.i.d. } \mathcal{N}(0,\Sigma_m),
\) and let the target responses be
\(
\mathbf{y}_0 \;=\; \sum_{m=1}^M \bX_0^{(m)}\bm\beta^{(m)} \;+\; \bm\varepsilon_0\), with \(\bm\varepsilon_0\sim\mathcal{N}(0,\sigma_\varepsilon^2 \bm I_{n_0})\). Next, we construct the shared activities across the target and the sources. For each modality \(m\), we select a shared subset \(S_m\subset A_m\) with \(|S_m|=I\). Then, for each source \(s\in\{1,\dots,K\}\), we draw its observed modality set \( \mathcal{M}_s \subset \{1,\dots,M\} \) uniformly at random with \(|\mathcal{M}_s|=L\), and for each \(m\in\mathcal{M}_s\), we generate a design matrix
\(\bX_s^{(m)}\in\mathbb{R}^{n_k\times p_m}\) with rows i.i.d. \(\mathcal{N}(0,\bm\Sigma_m)\).

\begin{figure}[h]
    \centering
    \includegraphics[width=0.49\linewidth]{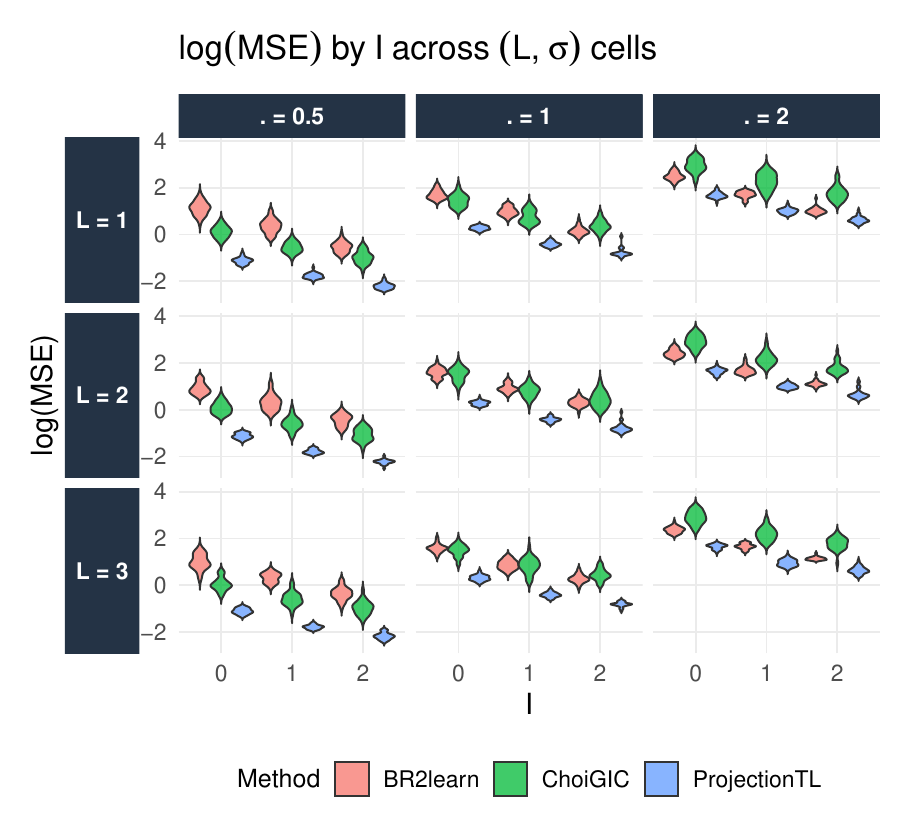}
    \includegraphics[width=0.49\linewidth]{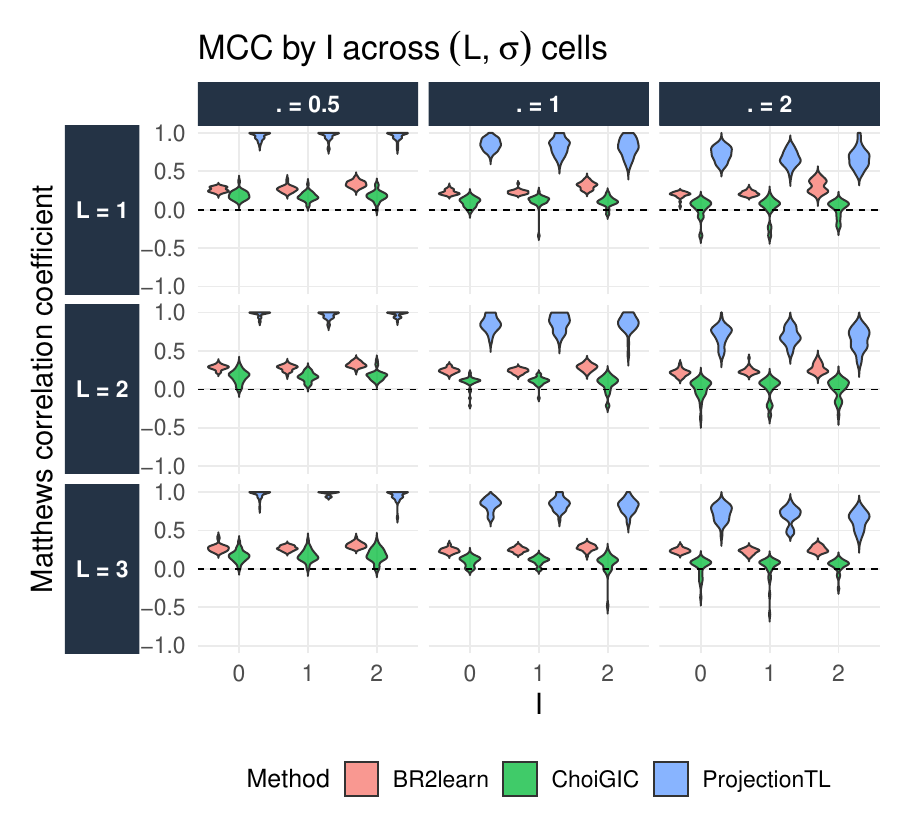}
    \caption{Results averaged over 200 replicates for $M=4$ modalities with $p_m = 40$ features per modality and $K=8$ sources where each source observes only  $L \in \{1,2,3\}$ modalities. The plots report mean MSE (left) and MCC (right) across combinations of noise level $\sigma_{\varepsilon} \in \{0.5,1,2\}$ and number of shared active coordinates $I \in \{0,1,2\}$.}
    \label{fig:multi-modal}
\end{figure}

Because this setting involves structured modality-level block missingness, many transfer-learning baselines used in earlier simulations are not directly applicable without ad hoc imputation or feature alignment. We compare ProjectionTL with BR2learn, a representative multi-task learning method designed for multi-modal data with block-wise missingness \citep{xu2025representation}. While BR2learn was originally proposed for joint multi-task estimation rather than transfer learning, we implement it in a target-evaluation setting by leveraging the target data for testing, providing a meaningful baseline for comparison under structured block missingness.

Figure~\ref{fig:multi-modal} compares ProjectionTL with BR2learn under modality-level block missingness across varying noise levels $\sigma_\varepsilon$, numbers of observed modalities $L$, and numbers of shared active coordinates $I$. ProjectionTL consistently achieves lower MSE than BR2learn when at least one informative coordinate is shared ($I \ge 1$), with the performance gap widening as the number of shared signals increases and as sources observe more modalities. The advantage is most pronounced in low-to-moderate noise regimes ($\sigma_\varepsilon = 0.5, 1$), where selective component-level borrowing allows ProjectionTL to effectively exploit partial modality overlap. When no shared signal is present ($I = 0$) or noise is high ($\sigma_\varepsilon = 2$), the two methods perform comparably, indicating that ProjectionTL avoids negative transfer and degrades gracefully. Overall, these results demonstrate that ProjectionTL reliably leverages informative cross-modal overlap while remaining robust under block missingness.

\section{Real Data Analysis Using ADNI}\label{sec:real_data}
We evaluate ProjectionTL using data from the Alzheimer's Disease Neuroimaging Initiative (ADNI), a large longitudinal study spanning multiple phases (ADNI1, ADNI-GO, ADNI2, ADNI3, and ADNI4) that has become a widely used resource for Alzheimer's disease research \citep{mueller2005alzheimer,weiner2017alzheimer}. Successive study phases differ in enrollment period, imaging protocols, and participant characteristics, creating a natural transfer learning setting in which earlier phases provide partially relevant but heterogeneous information for later phases.
Our goal is to model the relationship between regional amyloid burden and clinical dementia severity. The response variable is the baseline Clinical Dementia Rating-Sum of Boxes (CDRSB) score, a composite measure of cognitive and functional impairment, with higher scores indicating greater dementia severity. Predictors consist of $p=166$ regional standardized uptake value ratios (SUVRs) derived from amyloid PET imaging. We formulate the analysis as a transfer learning problem with one target cohort and multiple source cohorts corresponding to different ADNI phases.

This setting illustrates the challenges motivating ProjectionTL. Individual ADNI phases are relatively small compared with the high-dimensional imaging feature space, making target-only estimation unstable. At the same time, earlier phases contain potentially informative but heterogeneous relationships between imaging biomarkers and clinical dementia severity due to differences in acquisition protocols, study design, and patient populations. The objective is therefore to selectively borrow information from earlier phases while remaining robust to cross-phase heterogeneity.
We consider two transfer learning scenarios. The first investigates heterogeneous-source transfer using amyloid PET features available across study phases. The second considers multi-modal transfer with block-wise missingness, where MRI, amyloid PET, and gene expression measurements are not jointly observed for all participants.

\subsection{Study 1: Heterogeneous Sources with Single Modality}
We evaluate ProjectionTL in a heterogeneous-source transfer setting using amyloid PET imaging. The target cohort consists of newly enrolled participants from ADNI3 with complete baseline amyloid PET and Clinical Dementia Rating-Sum of Boxes (CDRSB) measurements. We excluded roll-over participants from earlier ADNI phases to avoid overlap between the target and source datasets. The predictor matrix contains $p=166$ regional amyloid PET standardized uptake value ratios (SUVRs), and the response is the baseline CDRSB score.
Two earlier ADNI phases serve as auxiliary sources. The first source comprises ADNI-GO participants ($n_1=128$), the earliest ADNI phase with systematic amyloid PET acquisition. The second source consists of ADNI2 participants ($n_2=769$), which expanded enrollment and standardized PET acquisition across a broader diagnostic spectrum. Although ADNI2 is generally more similar to ADNI3 than ADNI-GO, both source cohorts differ from the target in imaging protocols, enrollment criteria, and participant composition, inducing heterogeneous but related associations between amyloid burden and clinical dementia severity.
This multi-phase structure provides a realistic transfer learning problem in which earlier cohorts contain partially transferable information for estimating the regression model in the smaller ADNI3 target cohort. Four major sources of cross-phase heterogeneity are summarized in Supplement E.1.

\subsection{Study 2: Multi-modal Sources}
{We next evaluate ProjectionTL under block-wise missing modalities using the
ADNI2 cohort at the 48-month follow-up visit. The response is the CDRSB, a measure of cognitive and functional
impairment. Three complementary predictor modalities are considered:
(1) structural MRI, capturing neuroanatomical changes associated with
neurodegeneration; (2) amyloid PET, measuring cerebral amyloid burden; and
(3) gene expression, providing molecular information potentially associated
with disease progression. Subjects were linked across modalities using their
participant identifiers (PTID), including those with incomplete modality
measurements.}

\begin{figure}
    \centering
    \includegraphics[width=0.8\linewidth]{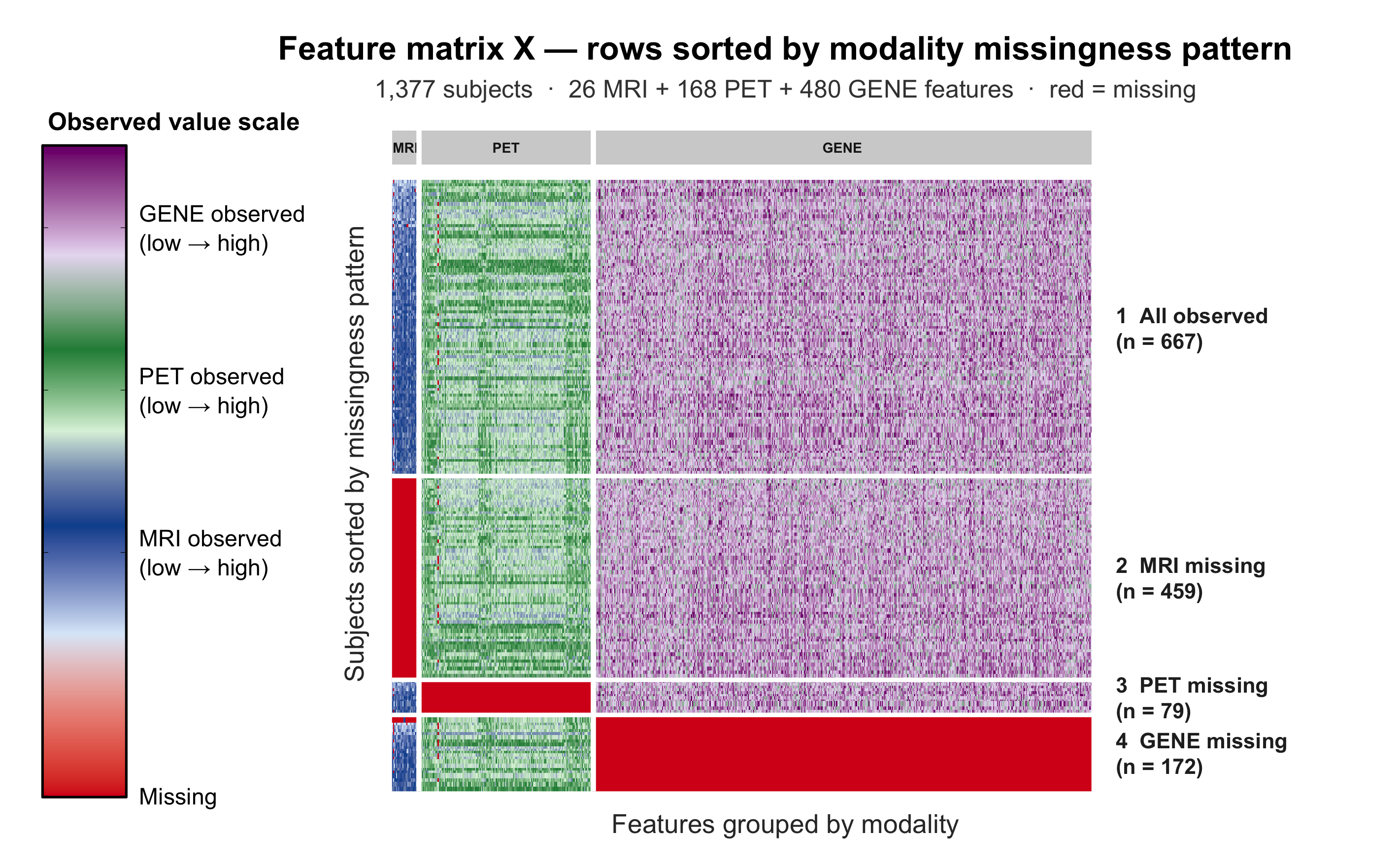}
    \caption{Schematic of the ADNI multi-modal transfer-learning design.
The target consists of all participants with complete MRI, PET and gene-expression measurements. The three auxiliary sources are the participants missing MRI, PET or gene expression; each source uses only its observed modalities.}
    \label{fig:heatmap_modalities}
\end{figure}

{Subjects were classified into four mutually exclusive groups according to
modality availability: (1) all three modalities observed, (2) MRI missing,
(3) PET missing, and (4) gene expression missing
(Figure~\ref{fig:heatmap_modalities}). Subjects with other missingness patterns
were excluded. The complete-case group formed the target dataset and contained
\(n_0=667\) participants; it was randomly divided into training (60\%),
validation (20\%), and test (20\%) subsets within each replication. The
MRI-missing, PET-missing, and gene-expression-missing groups formed
the imperfect three sources, respectively, with retained cohort sizes $(n_1,n_2,n_3)=(172,79,459).$
Thus, the target observes all three modalities, whereas each auxiliary source
lacks one modality block.}

{Because gene expression is high-dimensional, Sure Independence Screening
(SIS) \citep{fan2008sure} was performed separately within each random split
using only the target training data. Genes with fewer than ten usable target-training observations or near-zero target-training variance were excluded, and the remaining genes were ranked by the absolute values of their
marginal correlations with CDRSB in the target training sample. We retained the top $d=\left\lfloor\frac{n_{\mathrm{tr}}}{\log(n_{\mathrm{tr}})}\right\rfloor$
genes, with \(d\) capped by the number of eligible genes. The resulting gene
set was held fixed when fitting the auxiliary sources, tuning on the target
validation sample, and evaluating prediction on the target test sample;
neither validation nor test outcomes were used during screening. ProjectionTL
fit each incomplete source using only its observed modality blocks, and
unavailable source covariates were treated as structurally unobserved rather
than as observed zeros. Predictive performance was evaluated using the test
mean squared error for CDRSB.}

\subsection{Real-data Results}
\label{subsec:real_data_results}
\begin{figure}
    \centering
    \includegraphics[width=0.8\linewidth]{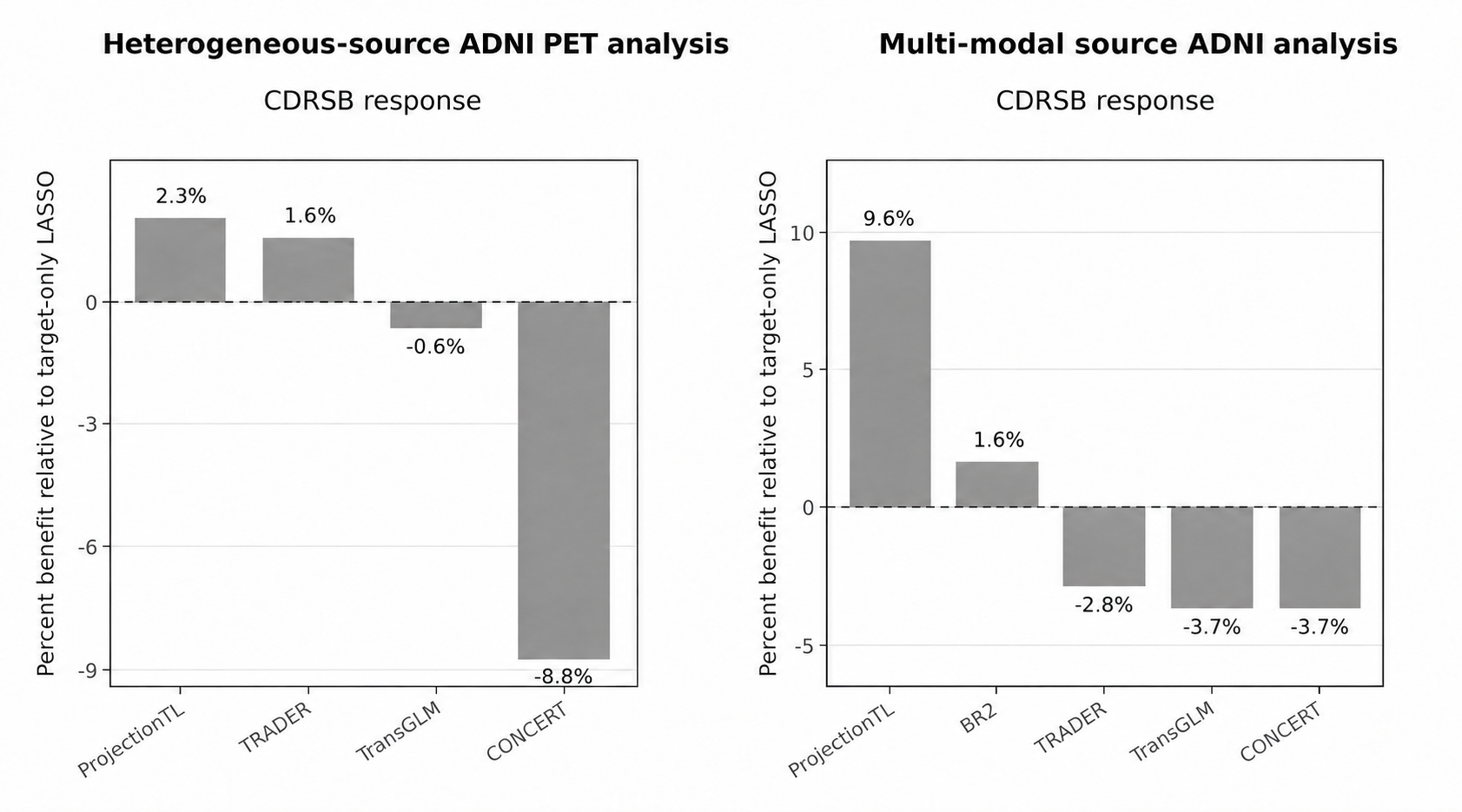}
    \caption{Percent benefit relative to target-only LASSO for CDRSB prediction in heterogeneous-source and multi-modal ADNI analyses.}
    \label{fig:realdata_benefit_lasso}
\end{figure}

Prediction performance was assessed by test-set
mean squared error (MSE) for the CDRSB cognitive outcome.
Figure~\ref{fig:realdata_benefit_lasso} reports the percentage reduction
in MSE relative to the target-only LASSO baseline,
$100\times(\mathrm{MSE}_{\mathrm{target\text{-}LASSO}} -
\mathrm{MSE}_{\mathrm{method}})/\mathrm{MSE}_{\mathrm{target\text{-}LASSO}}$,
so that positive values indicate improvement and negative values indicate
degradation relative to the target-only reference. Actual prediction-MSE figures are presented in Table~\ref{tab:my-table}.

\begin{table}[htbp]
\centering
\caption{Comparison of methods across two study settings. Values are reported as mean test MSE (SD).}
\label{tab:my-table}
\resizebox{\columnwidth}{!}{%
\begin{tabular}{lcccccc}
\toprule
\textbf{Setting} & \textbf{Target LASSO} & \textbf{BR2}
  & \cellcolor{gray!15}\textbf{ProjectionTL}
  & \textbf{TRADER} & \textbf{TransGLM} & \textbf{CONCERT} \\
\midrule
1. Heterogeneous sources
  & 1.622 (0.055) & --
  & \cellcolor{gray!15}1.586 (0.056)
  & 1.596 (0.056) & 1.632 (0.058) & 1.765 (0.063) \\
2. Multi-modal sources
  & 3.108 (0.161) & 3.059 (0.233)
  & \cellcolor{gray!15}2.811 (0.206)
  & 3.197 (0.135) & \textcolor{red}{3.222 (0.127)} & \textcolor{red}{3.225 (0.151)} \\
\bottomrule
\end{tabular}%
}
\end{table}


\textbf{Study 1:} When all source cohorts share the full PET feature space with the target,
ProjectionTL and TRADER are the only methods that improve over target-only
LASSO, with reductions of 2.3\% and 1.6\%, respectively. TransGLM is
essentially neutral ($-$0.6\%), and CONCERT degrades substantially
($-$8.8\%). The pattern suggests that in this setting the source signal is
modest but genuine, and that methods with stronger source-selection or
shrinkage mechanisms, including ProjectionTL's discrepancy-based projection and
TRADER's discrepancy-weighted ridge, are better able to isolate the useful
portion of the transferred information without overfitting to spurious
source variation. CONCERT's large degradation indicates that its
consensus-based aggregation, which pools all sources without explicit
source-quality screening, is susceptible to negative transfer when heterogeneity is high.

\textbf{Study 2:}
The multi-modal analysis presents a qualitatively different comparison.
{Source cohorts each lack one modality block, so standard transfer-learning methods that
require a common observed feature space, such as TRADER, TransGLM and CONCERT, cannot be applied
directly. We therefore pair each with an imputation step. For TransGLM and CONCERT, which use
individual-level source data, the missing block of each source is imputed by regression on the
source's observed blocks, with the imputation model fitted on the target training covariates; for
TRADER, which uses source coefficient estimates, each source's missing block of coefficients is
replaced by the average of the corresponding coefficients from the sources that observe that block.
Implementation details are given in Supplement~E.2.} All three methods show negative transfer relative to target-only LASSO ($-$3.7\%, $-$2.8\%, and $-$3.7\% for CONCERT, TRADER, and TransGLM, respectively). BR2, which accommodates reduced-form source information, achieves a modest gain of 1.6\%. ProjectionTL, the only method specifically designed for partially overlapping modalities, yields the largest improvement at 9.6\%,
using its source-calibration and coordinate-level borrowing steps to extract signal from sources that standard competitors must discard entirely.


\vspace{-1em}
\section{Discussion}\label{sec:discussion}
This paper proposes ProjectionTL, a two-stage framework that combines source-guided Bayesian priors with projection-based sparsification to address heterogeneous multi-source transfer learning. The central contribution lies in decoupling where to borrow (through source-level weighting via a Dirichlet prior) from what to borrow (through component-wise projection), enabling adaptive and interpretable transfer under partial overlap and block-wise missingness. This separation is conceptually clean and practically effective, as reflected in improved estimation accuracy and support recovery across simulations and real data analyses.
Empirically, the simulation results demonstrate that ProjectionTL achieves consistent improvements in estimation error and support recovery, particularly in regimes with partial overlap and heterogeneous source quality. The method performs especially well in intermediate regimes where naive pooling leads to over-borrowing and target-only estimation suffers from high variance. Importantly, the results also show that the method does not incur performance loss when no informative sources are present, indicating effective control of negative transfer. The real data analysis further illustrates the method’s applicability in complex biomedical settings, where data are collected across multiple phases with varying protocols and partially overlapping features.

Several aspects of the framework highlight opportunities for further refinement and extension. The projection step induces sparsity through $l_1$ regularization, which provides stability and interpretability in high-dimensional settings, though alternative projection schemes or debiasing strategies could further enhance estimation of large signals. While the method is built on a Bayesian formulation, inference is ultimately summarized through projection-based quantities such as inclusion frequencies. Therefore, developing formal post-projection uncertainty quantification would be a valuable direction for future work. In addition, the current formulation focuses on linear models, which serve as a foundational setting for high-dimensional transfer learning. Extending the framework to accommodate nonlinear relationships and more complex structures would broaden its applicability in modern data.

\section*{Data Availability Statement}\label{data-availability-statement}

Data used in the preparation of this article were obtained from the ADNI database (\url{http://adni.loni.usc.edu/}). As such, the investigators within ADNI contributed to the design and implementation of ADNI and/or provided data, but did not participate in the analysis or writing of this report. A complete listing of ADNI investigators can be found at: \url{http://adni.loni.usc.edu/wp-content/uploads/how_to_apply/ADNI_Acknowledgement_List.pdf}.

\vspace*{-1em}
{\setlength{\bibsep}{0pt plus 0.3ex}
\onehalfspacing
{
\nocite{zhang2023survey}
\bibliography{ref}

@article{pal2026improving,
  title={Improving {RCT}-Based {CATE} Estimation Under Covariate Mismatch via Double Calibration},
  author={Pal, Samhita and Huling, Jared D and Asiaee, Amir},
  journal={arXiv preprint arXiv:2603.17066},
  year={2026}
}

@article{li2023targeting,
  title={Targeting underrepresented populations in precision medicine: A federated transfer learning approach},
  author={Li, Sai and Cai, Tianxi and Duan, Rui},
  journal={The Annals of Applied Statistics},
  volume={17},
  number={4},
  pages={2970},
  year={2023}
}

@article{duan2022heterogeneity,
  title={Heterogeneity-aware and communication-efficient distributed statistical inference},
  author={Duan, Rui and Ning, Yang and Chen, Yong},
  journal={Biometrika},
  volume={109},
  number={1},
  pages={67--83},
  year={2022},
  publisher={Oxford University Press}
}

@article{li2024estimation,
  title={Estimation and inference for high-dimensional generalized linear models with knowledge transfer},
  author={Li, Sai and Zhang, Linjun and Cai, T Tony and Li, Hongzhe},
  journal={Journal of the American Statistical Association},
  volume={119},
  number={546},
  pages={1274--1285},
  year={2024},
  publisher={Taylor \& Francis}
}

@article{gu2025connections,
  title={On the connections among three transfer learning paradigms},
  author={Gu, Tian and Li, Sijia and Duan, Rui},
  journal={Stat},
  volume={14},
  number={4},
  pages={e70100},
  year={2025},
  publisher={Wiley Online Library}
}

@article{gu2023commute,
  title={COMMUTE: Communication-efficient transfer learning for multi-site risk prediction},
  author={Gu, Tian and Lee, Phil H and Duan, Rui},
  journal={Journal of biomedical informatics},
  volume={137},
  pages={104243},
  year={2023},
  publisher={Elsevier}
}

@article{bastani2021predicting,
  title={Predicting with proxies: Transfer learning in high dimension},
  author={Bastani, Hamsa},
  journal={Management Science},
  volume={67},
  number={5},
  pages={2964--2984},
  year={2021},
  publisher={INFORMS}
}

@article{lu2024enhancing,
  title={Enhancing genetic risk prediction through federated semi-supervised transfer learning with inaccurate electronic health record data},
  author={Lu, Yuying and Gu, Tian and Duan, Rui},
  journal={Statistics in biosciences},
  pages={1--22},
  year={2024},
  publisher={Springer}
}

@article{wang2023posterior,
  title={Posterior contraction and testing for multivariate isotonic regression},
  author={Wang, Kang and Ghosal, Subhashis},
  journal={Electronic Journal of Statistics},
  volume={17},
  number={1},
  pages={798--822},
  year={2023},
  publisher={The Institute of Mathematical Statistics and the Bernoulli Society}
}

@article{wang2023coverage,
  title={Coverage of credible intervals in bayesian multivariate isotonic regression},
  author={Wang, Kang and Ghosal, Subhashis},
  journal={The Annals of Statistics},
  volume={51},
  number={3},
  pages={1376--1400},
  year={2023},
  publisher={Institute of Mathematical Statistics}
}

@article{chakraborty2022rates,
  title={Rates and coverage for monotone densities using projection-posterior},
  author={Chakraborty, Moumita and Ghosal, Subhashis},
  journal={Bernoulli},
  volume={28},
  number={2},
  pages={1093--1119},
  year={2022},
  publisher={Bernoulli Society for Mathematical Statistics and Probability}
}

@article{yu2020optimal,
  title={Optimal sparse linear prediction for block-missing multi-modality data without imputation},
  author={Yu, Guan and Li, Quefeng and Shen, Dinggang and Liu, Yufeng},
  journal={Journal of the American Statistical Association},
  volume={115},
  number={531},
  pages={1406--1419},
  year={2020},
  publisher={Taylor \& Francis}
}

@article{zhang2023survey,
  title   = {A Survey on Negative Transfer},
  author  = {Zhang, Wen and Deng, Lingfei and Zhang, Lei and Wu, Dongrui},
  journal = {IEEE/CAA Journal of Automatica Sinica},
  volume  = {10},
  number  = {2},
  pages   = {305--329},
  year    = {2023},
  doi     = {10.1109/JAS.2022.106004}
}

@article{choi2024robust,
  title={Robust data integration from multiple external sources for generalized linear models with binary outcomes},
  author={Choi, Kyuseong and Taylor, Jeremy MG and Han, Peisong},
  journal={Biometrics},
  volume={80},
  number={1},
  pages={ujad005},
  year={2024},
  publisher={Oxford University Press}
}

@article{jin2021unit,
  author  = {Jin, Huaqing and Yin, Guosheng},
  title   = {Unit information prior for adaptive information borrowing from multiple historical datasets},
  journal = {Statistics in Medicine},
  year    = {2021},
  volume  = {40},
  number  = {25},
  pages   = {5657--5672},
  doi     = {10.1002/sim.9146}
}

@article{zhang2023unit,
  author  = {Zhang, Hongjian and Yin, Guosheng},
  title   = {Unit information prior for incorporating real-world evidence into randomized controlled trials},
  journal = {Statistical Methods in Medical Research},
  year    = {2023},
  volume  = {32},
  number  = {2},
  pages   = {229--241},
  doi     = {10.1177/09622802221143907}
}

@article{lai2026bayesian,
  author  = {Lai, Daoyuan and Madrid Padilla, Oscar Hernan and Gu, Tian},
  title   = {Bayesian transfer learning for enhanced estimation and inference},
  journal = {Journal of the American Statistical Association},
  year    = {2026},
  note    = {Accepted},
  url={https://arxiv.org/pdf/2412.02986},
}

@article{gili2026semiparametric,
  title={Semiparametric Bernstein--von Mises phenomenon via Isotonized Posterior in Wicksell’s problem},
  author={Gili, Francesco and Jongbloed, Geurt and Van Der Vaart, Aad},
  journal={The Annals of Statistics},
  volume={54},
  number={1},
  pages={383--407},
  year={2026},
  publisher={Institute of Mathematical Statistics}
}

@article{fan2008sure,
  title={Sure independence screening for ultrahigh dimensional feature space},
  author={Fan, Jianqing and Lv, Jinchi},
  journal={Journal of the Royal Statistical Society Series B: Statistical Methodology},
  volume={70},
  number={5},
  pages={849--911},
  year={2008},
  publisher={Oxford University Press}
}

@article{pal2025projection,
  title={Projection-posterior for variable selection: Weak limit and coverage},
  author={Pal, Samhita and Ghosal, Subhashis},
  journal={Electronic Journal of Statistics},
  volume={19},
  number={2},
  pages={3743--3770},
  year={2025},
  publisher={The Institute of Mathematical Statistics and the Bernoulli Society}
}

@article{asiaee2026improving,
  title={Improving {RCT}-Based Treatment Effect Estimation Under Covariate Mismatch via Calibrated Alignment},
  author={Asiaee, Amir and Pal, Samhita},
  journal={arXiv preprint arXiv:2603.19186},
  year={2026}
}

@article{asiaee2026b,
  title={B-CALM: Bias-Limited Bayesian Borrowing for RCT-Anchored Treatment Effects under Covariate Mismatch},
  author={Asiaee, Amir and Pal, Samhita},
  journal={arXiv preprint arXiv:2607.04036},
  year={2026}
}

@article{ghosh2026demographic,
  title={Demographic distribution matching between real-world and virtual phantom population},
  author={Ghosh, Dhrubajyoti and Tushar, Fakrul and Dahal, Lavsen and Vancoillie, Liesbeth and Lafata, Kyle J and Samei, Ehsan and Lo, Joseph Y and Luo, Sheng},
  journal={Medical Physics},
  volume={53},
  number={3},
  pages={e70364},
  year={2026},
  publisher={Wiley Online Library}
}

@article{weiner2017alzheimer,
  title={The Alzheimer's Disease Neuroimaging Initiative 3: Continued innovation for clinical trial improvement},
  author={Weiner, Michael W and Veitch, Dallas P and Aisen, Paul S and Beckett, Laurel A and Cairns, Nigel J and Green, Robert C and Harvey, Danielle and Jack Jr, Clifford R and Jagust, William and Morris, John C and others},
  journal={Alzheimer's \& Dementia},
  volume={13},
  number={5},
  pages={561--571},
  year={2017},
  publisher={Elsevier}
}

@article{mueller2005alzheimer,
  title={The Alzheimer's disease neuroimaging initiative},
  author={Mueller, Susanne G and Weiner, Michael W and Thal, Leon J and Petersen, Ronald C and Jack, Clifford and Jagust, William and Trojanowski, John Q and Toga, Arthur W and Beckett, Laurel},
  journal={Neuroimaging Clinics},
  volume={15},
  number={4},
  pages={869--877},
  year={2005},
  publisher={Elsevier}
}

@article{zhang2024covariate,
  title={Covariate-Elaborated Robust Partial Information Transfer with Conditional Spike-and-Slab Prior},
  author={Zhang, Ruqian and Zhang, Yijiao and Qu, Annie and Zhu, Zhongyi and Shen, Juan},
  journal={arXiv preprint arXiv:2404.03764},
  year={2024}
}

@article{xu2025representation,
  title={Representation retrieval learning for heterogeneous data integration},
  author={Xu, Qi and Qu, Annie},
  journal={arXiv preprint arXiv:2503.09494},
  year={2025}
}

@article{tian2023transfer,
  title={Transfer learning under high-dimensional generalized linear models},
  author={Tian, Ye and Feng, Yang},
  journal={Journal of the American Statistical Association},
  volume={118},
  number={544},
  pages={2684--2697},
  year={2023},
  publisher={Taylor \& Francis}
}

@article{li2022transfer,
  title={Transfer learning for high-dimensional linear regression: Prediction, estimation and minimax optimality},
  author={Li, Sai and Cai, T Tony and Li, Hongzhe},
  journal={Journal of the Royal Statistical Society Series B: Statistical Methodology},
  volume={84},
  number={1},
  pages={149--173},
  year={2022},
  publisher={Oxford University Press}
}

@article{wang2024multiresponse,
  title={Multi-response Regression for Block-missing Multi-modal Data without Imputation},
  author={Wang, Haodong and Li, Quefeng and Liu, Yufeng},
  journal={Statistica Sinica},
  volume={34},
  number={2},
  pages={527--546},
  year={2024},
  doi={10.5705/ss.202021.0170}
}

@article{xue2025statistical,
  title={Statistical Inference for High-Dimensional Linear Regression with Blockwise Missing Data},
  author={Xue, Fei and Ma, Rong and Li, Hongzhe},
  journal={Statistica Sinica},
  volume={35},
  number={1},
  pages={431--456},
  year={2025},
  doi={10.5705/ss.202022.0104}
}

@misc{wang2026multimodal,
  title={Multimodal Domain Adaptation under Label Shift and Blockwise Missing Modalities},
  author={Wang, Zebin and Dou, Ziang and Liu, Molei and Cai, Tianxi},
  year={2026},
  eprint={2608.01282},
  archivePrefix={arXiv},
  primaryClass={stat.ME},
  url={https://arxiv.org/abs/2608.01282}
}

@article{cai2024transfer,
  title={Transfer learning for functional mean estimation: Phase transition and adaptive algorithms},
  author={Cai, T Tony and Kim, Dongwoo and Pu, Hongming},
  journal={The Annals of Statistics},
  volume={52},
  number={2},
  pages={654--678},
  year={2024},
  publisher={Institute of Mathematical Statistics}
}

@article{pal2024bayesian,
  title={Bayesian High-dimensional Linear Regression with Sparse Projection-posterior},
  author={Pal, Samhita and Ghoshal, Subhashis},
  journal={arXiv preprint arXiv:2410.16577},
  year={2024}
}

@article{pal2024group,
  title={Bayesian High-dimensional Grouped-regression using Sparse Projection-posterior},
  author={Pal, Samhita and Ghosal, Subhashis},
  journal={arXiv preprint arXiv:2411.15713},
  year={2024}
}

@article{gu2024robust,
  title={Robust angle-based transfer learning in high dimensions},
  author={Gu, Tian and Han, Yi and Duan, Rui},
  journal={Journal of the Royal Statistical Society Series B: Statistical Methodology},
  pages={qkae111},
  year={2025},
  publisher={Oxford University Press UK}
}

@article{chang2024heterogeneous,
  title={Heterogeneous transfer learning for high dimensional regression with feature mismatch},
  author={Chang, Jae Ho and Russo, Massimiliano and Paul, Subhadeep},
  journal={arXiv preprint arXiv:2412.18081},
  year={2024}
}

@article{li2024adaptive,
  title={Adaptive and efficient learning with blockwise missing and semi-supervised data},
  author={Li, Yiming and Yang, Xuehan and Wei, Ying and Liu, Molei},
  journal={arXiv preprint arXiv:2405.18722},
  year={2024}
}

@article{zhao2023heterogeneous,
  title={Heterogeneous transfer learning for building high-dimensional generalized linear models with disparate datasets},
  author={Zhao, Ruzhang and Kundu, Prosenjit and Saha, Arkajyoti and Chatterjee, Nilanjan},
  journal={arXiv preprint arXiv:2312.12786},
  year={2023}
}
}}
%


\end{document}